\documentclass{article}

\usepackage[final]{neurips_data_2024}

\usepackage[utf8]{inputenc} 
\usepackage[T1]{fontenc}    
\usepackage{hyperref}       
\hypersetup{
    colorlinks=true,
    linkcolor=blue,
    filecolor=magenta,      
    urlcolor=magenta,
    citecolor=blue,
}
\usepackage{url}            
\usepackage{booktabs}       
\usepackage{amsfonts}       
\usepackage{nicefrac}       
\usepackage{microtype}      
\usepackage{xcolor}         

\usepackage{natbib}
\usepackage{amsmath}
\usepackage{multirow}
\usepackage{threeparttable}
\usepackage{wrapfig}
\usepackage{CJK}
\usepackage{array}
\usepackage{caption}        
\usepackage{subcaption}
\usepackage{makecell}
\usepackage{rotating}
\usepackage{soul}
\usepackage{comment}
\usepackage{pifont}
\usepackage{enumitem,amssymb}
\usepackage{tcolorbox}
\usepackage{csquotes}
\usepackage{fontawesome}
\usepackage{siunitx}
\usepackage{amssymb}
\usepackage{pifont}

\renewcommand{\thefootnote}{\fnsymbol{footnote}}
\title{DiReCT: Diagnostic Reasoning for Clinical Notes \\ via Large Language Models}

\author{
\normalfont{\textbf{Bowen Wang}$^{\clubsuit\diamondsuit}$\footnote{Equal contribution.}, \textbf{Jiuyang Chang}$^{\heartsuit}$\footnotemark[1], \textbf{Yiming Qian}$^{\spadesuit}$\footnote{Corresponding author.}\hspace{0.15cm}, \textbf{Guoxin Chen}$^{\bigstar}$,} \textbf{Junhao Chen}$^{\diamondsuit}$, \\
\textbf{Zhouqiang Jiang}$^{\diamondsuit}$, \textbf{Jiahao Zhang}$^{\diamondsuit}$, \textbf{Yuta Nakashima}$^{\diamondsuit\clubsuit}$, \textbf{Hajime Nagahara}$^{\diamondsuit\clubsuit}$\\
$^{\clubsuit}$Premium Research Institute for Human Metaverse Medicine (WPI-PRIMe), Osaka University, \\
$^\heartsuit$Department of Cardiology, The First Affiliated Hospital of Dalian Medical University,\\
$^\bigstar$Institute of Computing Technology, Chinese Academy of Science \\
$^\diamondsuit$D3 Center, Osaka University,
$^\spadesuit$Agency for Science, Technology and Research (A*STAR), \\
\{\texttt{wang}, \texttt{n-yuta}, \texttt{nagahara}\}\texttt{@ids.osaka-u.ac.jp} \\
\texttt{changjiuyang@firsthosp-dmu.com}\\
\texttt{qiany@ihpc.a-star.edu.sg}, \texttt{gx.chen.chn@gmail.com}\\
\{\texttt{junhao}, \texttt{zhouqiang}, \texttt{jiahao}\}\texttt{@is.ids.osaka-u.ac.jp}\\}

\begin{document}

\maketitle
   
\footnotetext[1]{Equal contribution.}
\footnotetext[2]{Corresponding author.}

\begin{abstract}
Large language models (LLMs) have recently showcased remarkable capabilities, spanning a wide range of tasks and applications, including those in the medical domain. Models like GPT-4 excel in medical question answering but may face challenges in the lack of interpretability when handling complex tasks in real clinical settings. We thus introduce the diagnostic reasoning dataset for clinical notes (DiReCT), aiming at evaluating the reasoning ability and interpretability of LLMs compared to human doctors. It contains 511 clinical notes, each meticulously annotated by physicians, detailing the diagnostic reasoning process from observations in a clinical note to the final diagnosis. Additionally, a diagnostic knowledge graph is provided to offer essential knowledge for reasoning, which may not be covered in the training data of existing LLMs. Evaluations of leading LLMs on DiReCT bring out a significant gap between their reasoning ability and that of human doctors, highlighting the critical need for models that can reason effectively in real-world clinical scenarios \footnote{Data and code are available at https://github.com/wbw520/DiReCT.}.
\end{abstract}


\setcounter{footnote}{0}
\renewcommand{\thefootnote}{\arabic{footnote}}

\section{Introduction}\label{sec:intro}

Recent advancements of large language models (LLMs) \citep{LLMSurvey} have ushered in new possibilities and challenges for a wide range of natural language processing (NLP) tasks \citep{min2023recent}. In the medical domain, these models have demonstrated remarkable prowess \citep{anil2023palm,han2023medalpaca}, particularly in medical question answering (QA) \citep{jin2021disease}. Leading-edge models, such as GPT-4 \citep{GPT-4}, exhibit profound proficiency in understanding and generating text \citep{bubeck2023sparks}, even achieved high scores on the United States Medical Licensing Examination (USMLE) questions \citep{nori2023capabilities}. 

\begin{figure}[t]
\centering
\includegraphics[width=0.95\textwidth]{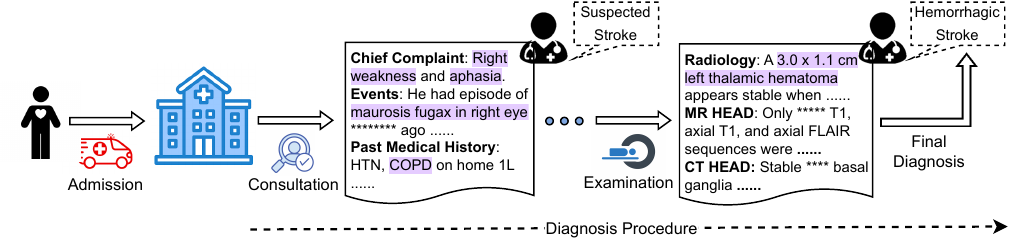}
\caption{When a patient is admitted, an initial consultation takes place to collect subjective information. Subsequent observations may then require further examination to confirm the diagnosis.}
\label{fig:1}
\end{figure}

Despite the advancements, interpretability is critical, particularly in medical NLP tasks \citep{lievin2024can} because these tasks directly impact patient health and treatment decisions. Without clear interpretability, there's a risk of misdiagnosis and improper treatment, making it vital for ensuring medical safety. Some studies assess this capability over medical QA \citep{pal2022medmcqa,li2023explaincpe,chen2024benchmarking} or natural language inference (NLI) \citep{jullien2023semeval}. Putting more attention on interpretability, they use relatively simple tasks as testbeds, taking short text as input. Nevertheless, real-world clinical tasks are often more complex \citep{gao2023leveraging}, as illustrated in Figure \ref{fig:1}, a typical diagnosis requires comprehending and combining various information, such as health records, physical examinations, and laboratory tests, for further reasoning of possible diseases in a step-by-step manner following the established guidelines. This observation suggests that both \textit{perception}, or reading (e.g., finding necessary information in the medical record) and \textit{reasoning} (determining the disease based on the observations) should be counted when evaluating interpretability in LLM-based medical NLP tasks.

For a more comprehensive evaluation of LLMs for supporting diagnosis in a more realistic setting, we propose a \textbf{\underline{Di}}agnostic \textbf{\underline{Re}}asoning dataset for \textbf{\underline{C}}linical no\textbf{\underline{T}}es (DiReCT). The task basically is predicting the diagnosis from a \textit{clinical note} of a patient, which is a collection of various medical records, written in natural language. 
Our dataset contains 511 clinical notes spanning 25 disease categories, sampled from a publicly available database, MIMIC-IV \citep{johnson2023mimic}. 
Each clinical note undergoes fine-grained annotation by professional physicians. The annotators (i.e., the physicians) are responsible for identifying the text, or the \textit{observation}, in the note that leads to a certain diagnosis, as well as the explanation. 
The dataset also provides a diagnostic knowledge graph based on existing diagnostic guidelines to facilitate more consistent annotations and to supply a model with essential knowledge for reasoning that might not be encompassed in its training data. 

To underscore the challenge offered by our dataset, we propose a simple AI-agent based baseline \citep{xi2023rise,tang2023medagents} that utilizes the knowledge graph to decompose the diagnosis into a sequence of diagnoses from a smaller number of observations. Our experimental findings indicate that current state-of-the-art LLMs still fall short of aligning well with human doctors.


\textbf{Contribution}. DiReCT offers a new challenge in diagnosis from a complex clinical note with explicit knowledge of established guidelines. This challenge aligns with a realistic medical scenario that doctors are experiencing. In the application aspect, the dataset facilitates the development of a model to support doctors in diagnosis, which is error-prone \citep{middleton2013enhancing,liu2022note}. From the technical aspect, the dataset can benchmark models' ability to read long text and find necessary observations for \textit{multi-evidence entailment tree} reasoning, an extension of the original entailment tree explanation \citep{dalvi-explaining} for complex scenarios in medical NLP tasks. As shown in Figure \ref{fig:annotated_HF}, this is not trivial because of the variations in writing; superficial matching does not help, and medical knowledge is vital. Meanwhile, reasoning itself is facilitated by the knowledge graph. The model does not necessarily have the knowledge of diagnostic guidelines. With this choice, the knowledge graph explains the reasoning process, which is also beneficial when deploying such a diagnosis assistant system in practical uses.

\section{Related Works}\label{sec:relate_works}
\textbf{Natural language explanation}. Recent advancements in NLP have led to significant achievements \citep{min2023recent}. However, existing models often lack explainability, posing potential risks \citep{danilevsky-survey,gurrapu2023rationalization}. Numerous efforts have been made to address this challenge. One effective approach is to provide a human-understandable \textit{plain text} explanation alongside the model's output \citep{camburu2018snli, rajani-explain}. Another strategy involves identifying \textit{evidence} within the input that serves as a rationale for the model's decisions, aligning with human reasoning \citep{deyoung-eraser}. Expanding on this concept, \citep{jhamtani2020learning} introduces chain-structured explanations, given that a diagnosis can demand multi-hop reasoning. This idea is further refined by ProofWriter \citep{tafjord2020proofwriter} through a proof stage for explanations, and by \citep{zhao-multi-step} through retrieval from a corpus. \citep{dalvi-explaining} proposes the \textit{entailment tree}, offering more detailed explanations and facilitating inspection of the model’s reasoning. More recently, \citep{zhang2024cumulative} employed cumulative reasoning to tap into the potential of LLMs to provide explanation via a \textit{directed acyclic graph}. Although substantial progress has been made, interpreting NLP tasks in medical domains remains an ongoing challenge \citep{lievin2024can}.

\textbf{Benchmarks of interpretability in the medical domain} Several datasets are designed to assess a model's reasoning together with its interpretability in medical NLP (Table \ref{tab:dataset-summary}). MedMCQA \citep{pal2022medmcqa} and other medical QA datasets \citep{li2023explaincpe,chen2024benchmarking} provide plain text as explanations for QA tasks. NLI4CT \citep{jullien2023semeval} uses clinical trial reports, focusing on NLI supported by multi-hop reasoning. N2N2 \citep{gao-etal-2022-hierarchical} proposes a summarization (Sum) task for a diagnosis based on multiple pieces of evidence in the input clinical note. NEJM CPC \citep{zack2023clinical} interprets clinicians' diagnostic reasoning as plain text for reasoning clinical diagnosis (CD). DR.BENCH \citep{gao2023dr} aggregates publicly available datasets to assess the diagnostic reasoning of LLMs. Utilizing an multi-evidence entailment tree explanation, DiReCT introduces a more rigorous task to assess whether LLMs can align with doctors' reasoning in real clinical settings.

\begin{table}[t]
\centering
\caption{Comparison of existing datasets for medical reasoning tasks and ours. ``t'' and ``w'' mean tokens and words for the length of input, respectively.}
\label{tab:dataset-summary}
\resizebox{1\textwidth}{!}{
\begin{tabular}{lccccc}
\toprule
\textbf{Dataset} &  \textbf{Task} & \textbf{Data Source} & \textbf{Length} & \textbf{Explanation} & \textbf{\# Cases} \\
\midrule
MedMCQA \citep{pal2022medmcqa} & QA & Examination & 9.93 t & Plain Text & 194,000   \\
ExplainCPE \citep{li2023explaincpe} & QA & Examination & 37.79 w & Plain Text & 7,000   \\
JAMA Challenge \citep{chen2024benchmarking} & QA & Clinical Cases & 371 w & Plain Text & 1,524   \\
Medbullets \citep{chen2024benchmarking} & QA & Online Questions & 163 w & Plain Text & 308  \\
N2N2 \citep{gao-etal-2022-hierarchical} & Sum & Clinical Notes & 785.46 t & Evidences & 768  \\
NLI4CT \citep{jullien2023semeval} & NLI & Clinical Trail Reports & 10-35 t & Multi-hop & 2,400 \\
NEJM CPC \citep{zack2023clinical} & CD & Clinical Cases & - & Plain Text & 2,525 \\
DiReCT (Ours) & CD & Clinical Notes & 1074.6 t & Entailment Tree & 511 \\
\bottomrule
\end{tabular}
}
\end{table}
\section{A benchmark for Clinical Notes Diagnosis}

This section first detail clinical notes (Section \ref{sec:data}). We also describes the knowledge graph that encodes existing guidelines (Section \ref{sec:flow}). Our task definition, which tasks a clinical note and the knowledge graph as input is given in Section \ref{sec:task}. We then present our annotation process for clinical notes (Section \ref{sec:annotation}) and the evaluation metrics (Section \ref{sec:evaluation}).

\subsection{Clinical Notes}\label{sec:data}

Clinical notes used in DiReCT are stored in the SOAP format \citep{weed1970medical}. A clinical note comprises four components: In the \textit{subjective} section, the physician records the patient's chief complaint, the history of present illness, and other subjective experiences reported by the patient. The \textit{objective} section contains structural data obtained through examinations (inspection, auscultation, etc.) and other measurable means. The \textit{assessment} section involves the physician's analysis and evaluation of the patient's condition. This may include a summary of current status, \textit{etc}. Finally, the \textit{plan} section outlines the physician's proposed treatment and management plan. This may include prescribed medications, recommended therapies, and further investigations. A clinical note also includes a primary discharge diagnosis (PDD) in the assessment section. 

DiReCT's clinical notes are sourced from the MIMIC-IV dataset \citep{johnson2023mimic} (PhysioNet Credentialed Health Data License 1.5.0), which encompasses over 40,000 patients admitted to the intensive care units. Each note contains clinical data for a patient. To construct DiReCT, we curated a subset of 511 notes whose PDDs fell within one of 25 disease categories $i$ in 5 medical domains. 

In our task, a note $R=\{r\}$ is an excerpt of 6 clinical data in the subjective and objective sections (i.e., $|R| = 6$): chief complaint, history of present illness, past medical history, family history, physical exam, and pertinent results.\footnote{We excluded data, such as review system and social history, because they are often missing in the original clinical notes and are less relevant to the diagnosis.} We also identified the PDD $d^\star$ associated with $R$.\footnote{All clinical notes in DiReCT are related to only one PDD, and there is no secondary discharge diagnosis.} The set of $d^\star$'s for all $R$'s collectively forms $\mathcal{D}^\star$. We manually removed any descriptions that disclose the PDD in $R$.

\subsection{Diagnostic Knowledge Graph}\label{sec:flow}

Existing knowledge graphs for the medical domain, e.g., UMLS KG \citep{bodenreider2004unified}, lack the ability to provide specific clinical decision support (e.g., diagnostic threshold, context-specific data, dosage information, etc.), which are critical for accurate diagnosis. 

Our knowledge graphs $\mathcal{K} = \{k_i\}$ is a collection of graph $k_i$ for disease category $i$.
$k_i$ is based on the diagnosis criteria in existing guidelines (refer to supplementary material for details). $k_i$'s nodes are either premise $p \in \mathcal{P}_i$ (medical statement, e.g., \texttt{Headache is a symptom of}) and diagnoses $d \in \mathcal{D}_i$ (e.g., \texttt{Suspected Stroke}). $k_i$ consists of two different types of edges. One is \textit{premise-to-diagnosis} edges $\mathcal{S}_i = \{(p, d)\}$; an edge is from $p$ to $d$. This edge represents the necessary premise $p$ to make a diagnosis $d$. We refer to them as \textit{supporting} edges. The other is \textit{diagnosis-to-diagnosis} edges $\mathcal{F}_i = \{(d, d')\}$, where $d, d' \in \mathcal{D}_i$ and the edge is from $d$ to $d'$, which represents the diagnostic flow. These edges are referred to as \textit{procedural} edges. 

A disease category is defined according to an existing guideline, which starts from a certain diagnosis; therefore, a procedural graph $g_i = (\mathcal{D}_i, \mathcal{F}_i)$ ($\mathcal{G}=\{g_i\}$) has only one root node and arbitrarily branches toward multiple leaf nodes that represent PDDs (i.e., the clinical notes in DiReCT are chosen to cover all leaf nodes of $g_i$). Thus, $g_i$ is a \textit{tree}. We denote the set of the leaf nodes (or PDDs) as $\mathcal{D}^\star_i \subset \mathcal{D}_i$. The knowledge graph is denoted by $k_i = (\mathcal{D}_i, \mathcal{P}_i, \mathcal{S}_i, \mathcal{F}_i)$.

\begin{wrapfigure}{R}{0.64\textwidth}
\centering
\vspace*{-4mm}
\includegraphics[width=0.64\textwidth]{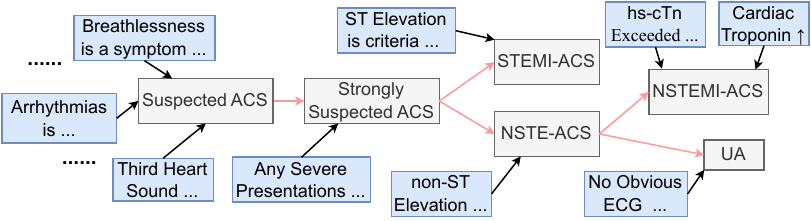} 
\caption{A part of $k_i$ for $i$ being \texttt{\small Acute Coronary Syndromes}.} 
\label{fig:procedure_HF}
\end{wrapfigure}

Figure \ref{fig:procedure_HF} shows a part of $k_i$, where $i$ is \texttt{Acute Coronary Syndromes (ACS)}. Premises in $\mathcal{P}_i$ and diagnoses in $\mathcal{D}_i$ are given in the blue and gray boxes, while PDDs in $\mathcal{D}^\star_i$ are ones without outgoing edges (i.e., \texttt{STEMI-ACS} and \texttt{NSTEMI-ACS}, and \texttt{UA}). The black and red arrows are edges in $\mathcal{S}$ and $\mathcal{F}$, respectively, where the black arrows indicate the supporting edges. 

$\mathcal{K}$ serves two essential functions: (1) They serve as the gold standard for annotation, guiding doctors in the precise and uniform interpretation of clinical notes. (2) Our task also allows a model to use them to ensure the output from an LLM can be closely aligned with the reasoning processes of medical professionals.

\subsection{Data Annotation} \label{sec:annotation}

\begin{figure}[t]
\centering
\includegraphics[width=.97\textwidth]{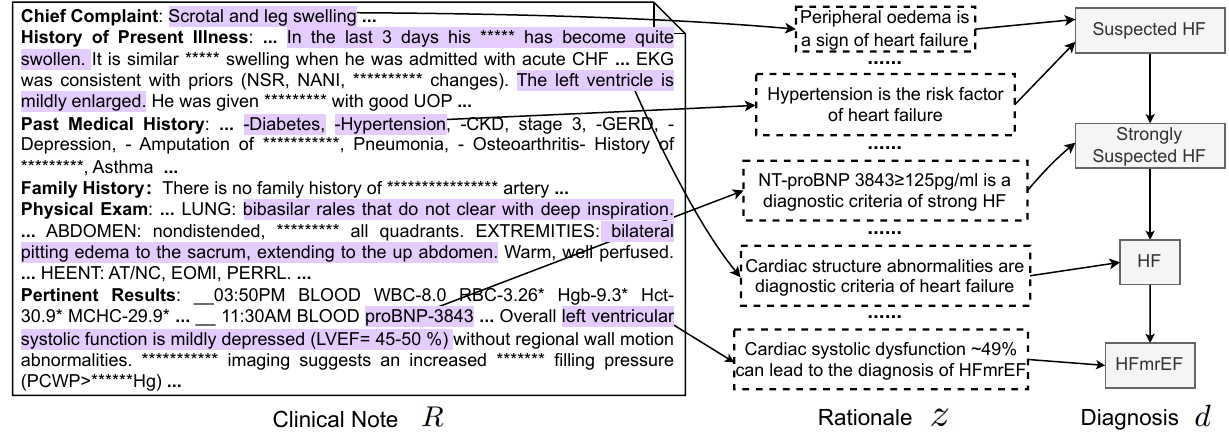}
\caption{An annotation sample of \texttt{Heart Failure} (\texttt{HF}). The left part is the clinical note alongside extracted observations by a doctor. The middle part outlines the steps of the rationale for the premise corresponding to each diagnostic node shown in the right part.}
\label{fig:annotated_HF}
\end{figure}

Let $d^\star \in \mathcal{D}^{\star}_i$ denote the PDD of disease category $i$ associated with $R$. We can find a subgraph $k_i(d^\star)$ of $k_i$ that contains all ancestors of $d^\star$, including premises in $\mathcal{P}_i$. We also denote the set of supporting edges in $k_i(d^\star)$ as $\mathcal{S}_i(d^\star)$. Our annotation process is, for each supporting edge $(p, d) \in \mathcal{S}_i(d^\star)$, to extract observation $o \in \mathcal{O}$ in $R$ (highlighted text in the clinical note in Figure \ref{fig:annotated_HF}) and provide rationalization $z$ of this \textit{deduction} why $o$ is a support for $d$ or corresponds to $p$.\footnote{All annotations strictly follow the procedural flow in $k_i$, and each observation is only related to one diagnostic node. If $R$ does not provide sufficient observations for the PDD (which may happen when a certain test is omitted), the annotators were asked to add plausible observations to $R$. Refer to amended data points in supplementary for details.} They form the explanation $\mathcal{E} = \{(o, z, d)\}$ for $(R, d^\star)$. This annotation process was carried out by 9 clinical physicians and subsequently verified for accuracy and completeness by three senior medical experts. 

\begin{wraptable}{R}{0.53\textwidth}
\centering
\footnotesize
\vspace*{-4mm}
\caption{Statistics of DiReCT.}
\label{table_statistic}
\resizebox{0.52\textwidth}{!}{%
\begin{tabular}{lcccccc}
\toprule
Medical domain & \# cat. & \# samples & $|\mathcal{D}_i|$ & $|\mathcal{D}^{\star}_i|$ & $|\mathcal{O}|$ & Length \\
\midrule
Cardiology & 7& 184 & 27 & 16 & 8.7 & 1156.6 t \\
Gastroenterology & 4 & 103 & 11 & 7 & 4.3 & 1026.0 t\\
Neurology & 5 & 77 & 17 & 11 & 11.9 & 1186.3 t\\
Pulmonology & 5 & 92 & 26 & 17 & 10.7 & 940.7 t\\
Endocrinology & 4 & 55 & 20 & 14 & 6.9 & 1063.5 t\\
\midrule
Overall & 25 & 511 & 101 & 65 & 8.5 & 1074.6 t\\
\bottomrule
\end{tabular}
}
\end{wraptable}
Table \ref{table_statistic} summarizes statistics of our dataset. The second and third columns (``\# cats.'' and ``\# samples'') show the numbers of disease categories and samples in the respective medical domains. $|\mathcal{D}_i|$ and $|\mathcal{D}_i^\star|$ are the total numbers of diagnoses (diseases) and PDDs, summed over all diagnostic categories in the medical domain, respectively. $|\mathcal{O}|$ is the average number of annotated observations. ``Length'' is the average number of tokens in $R$.

\subsection{Task Definition}\label{sec:task}

We propose two tasks with different levels of supplied external knowledge. The first task is, given $R$ and $\mathcal{G}$, to predict the associated PDD $d^\star$ and generate an explanation $\mathcal{E}$ that explains the model's diagnostic procedure from $R$ to $d^\star$, i.e., letting $M$ denote a model:
\begin{align}
    \hat{d}^\star, \hat{\mathcal{E}} = M(R, \mathcal{G}),
\end{align}
where $\hat{d}^\star \in \cup_i \mathcal{D}^\star_i$ and $\hat{\mathcal{E}}$ are predictions for the PDD and explanation, respectively. With this task, the knowledge of specific diagnostic procedures in existing guidelines can be used for prediction, facilitating interpretability. The second task takes $\mathcal{K}$ as input instead of $\mathcal{G}$, i.e.,:
\begin{align}
    \hat{d}^\star, \hat{\mathcal{E}} = M(R, \mathcal{K}).
\end{align}
This task allows for the use of broader knowledge of premises for prediction. One may also try a task without any external knowledge.


\subsection{Evaluation Metrics}
\label{sec:evaluation}

We designed three metrics to quantify the predictive performance over our benchmark. 

\noindent (1) \textit{Accuracy of diagnosis} $\textit{Acc}^\text{diag}$ evaluates if a model can correctly identify the diagnosis. $\textit{Acc}^\text{diag} = 1$ if $d^{\star} = \hat{d}$, and $\textit{Acc}^\text{diag} = 0$ otherwise. The average is reported. 

\noindent (2) \textit{Completeness of observations} $\textit{Obs}^\text{comp}$ evaluates whether a model extracts all and only necessary observations for the prediction. Let $\mathcal{O}$ and $\hat{\mathcal{O}}$ denote the sets of observations in $\mathcal{E}$ and $\hat{\mathcal{E}}$, respectively. The metric is defined as
$\textit{Obs}^\text{comp} = |\mathcal{O} \cap \hat{\mathcal{O}}| / |\mathcal{O} \cup \hat{\mathcal{O}}|$, where the numerator is the number of observations that are common in both $\mathcal{O}$ and $\hat{\mathcal{O}}$.\footnote{We find the common observations with an LLM (refer to the supplementary material for more detail).} This metric simultaneously evaluates the correctness of each observation and the coverage. To supplement it, we also report the precision $\textit{Obs}^\text{pre}$ and recall $\textit{Obs}^\text{rec}$, given by 
$\textit{Obs}^\text{pre} = |\mathcal{O} \cap \hat{\mathcal{O}}| / |\hat{\mathcal{O}}|$ and $\textit{Obs}^\text{rec} = |\mathcal{O} \cap \hat{\mathcal{O}}| / |\mathcal{O}|$. 

\noindent (3) \textit{Faithfulness of explanations} evaluates if the diagnostic flow toward the PDD is fully supported by observations with faithful rationalizations. 
This involves establishing a one-to-one correspondence between deductions in the prediction and the ground truth. We use the correspondences established for computing $\textit{Obs}^\text{comp}$. Let $o \in \mathcal{O}$ and $\hat{o} \in \hat{\mathcal{O}}$ denote corresponding observations. This correspondence is considered successful if $z$ and $\hat{z}$ as well as $d$ and $\hat{d}$ associated with $o$ and $\hat{o}$ matches. Let $m(\mathcal{E}, \hat{\mathcal{E}})$ denote the number of successful matches. We use the ratio of $m(\mathcal{E}, \hat{\mathcal{E}})$ to $|\mathcal{O} \cap \hat{\mathcal{O}}|$ and $|\mathcal{O} \cup \hat{\mathcal{O}}|$ as evaluation metrics $\textit{Exp}^\text{com}$ and $\textit{Exp}^\text{all}$, respectively, to see failures come from observations or explanations and diagnosis.


\section{Baseline}\label{sec:methods}

\begin{figure}[t]
\centering
\includegraphics[width=1\linewidth]{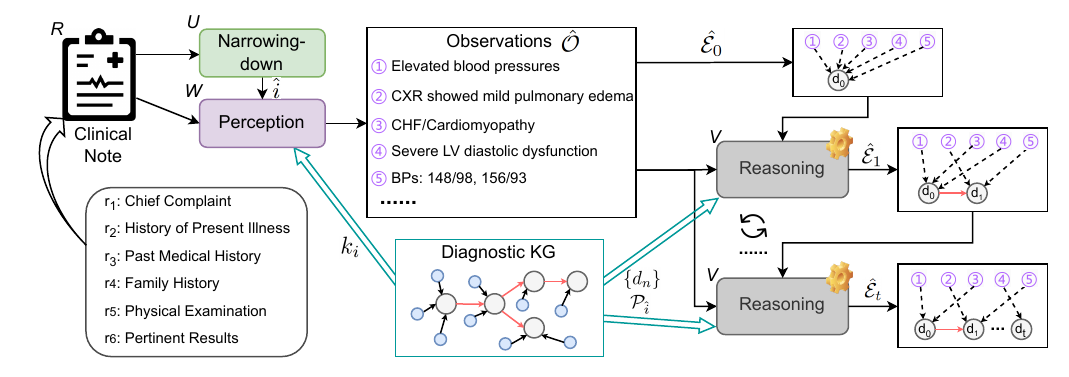}
\caption{Pipeline of our baseline. The dotted line in the right-most boxes means deductions from an observation to a diagnosis.}
\label{fig:task_samples}
\end{figure}

Figure \ref{fig:task_samples} provides an overview of our baseline, which comprises three LLM-based modules: narrowing-down ($U$), perception ($W$), and reasoning ($V$). In our experiments, each module utilizes the same type of LLM with different prompts (refer to the supplementary material for more details). $U$ analyze the entire note $R$ to determine the possible disease type $\hat{i}$. $W$ extracts observations that may lead to diseases from each $r$, producing a list of original disease descriptions. $V$ iteratively derives possible diseases from observations based on the diagnosis knowledge graph, providing rationales for each deduction $(o, z, d)$. 

The narrowing-down module $U$ takes $R$ as input to make a prediction $\hat{i}$ of the disease category, i.e., $\hat{i} = U(R)$. Let $d_t \in \mathcal{D}_{\hat{i}}$ be the diagnosis that has been reached with $t$ iterations over $k_{\hat{i}}$, where $t$ corresponds to the depth of node $d_t$ and so is less than or equal to the depth of $k_{\hat{i}}$. $d_0$ is the root node of $k_{\hat{i}}$. For $d_0$, we apply the perception module to extract all observations in $R$ and explanation $\mathcal{E}_0$ to support $d_0$ as
\begin{align}
    \hat{\mathcal{O}}, \hat{\mathcal{E}_0} = W(d_0, k_{\hat{i}}).
\end{align}
$k_{\hat{i}}$ is supplied to facilitate the model to extract all observations for the following reasoning process.\footnote{We used only pairs of an observation and a premise. We abuse $\mathcal{K}$ to mean this for notation simplicity. The perception model can also utilize $g_i$ instead of $k_i$ for the first task.} 
After the perception module $W$ (iteration $t = 0$), we obtain all observations $\mathcal{\hat{O}}$, the root node of the diagnosis $d_0$, and an explanation $\hat{\mathcal{E}}_0$ for the initial iteration. Assuming that by iteration $t $, we already know the diagnosis for iteration $t$ as $d_t$. $\{d_n\}$ is the set of $d_t$'s children, and $\mathcal{P}_{\hat{i}}(\{d_n\})$ represents the corresponding premises that support each $d_n$. $V$ identifies the diagnosis for the next step, $d_{t+1}$, and provides a justification $\mathcal{E}_{t+1}$. $V$ will verify if there is any $\hat{o}$ in $\mathcal{\hat{O}}$ that supports a $d_n$. If fully supported, $d_n$ is identified as $d_{t+1}$ for the $(t+1)$-th iteration, i.e., 
\begin{align}
d_{t+1}, \hat{\mathcal{E}}_{t+1} = V(\hat{\mathcal{O}}, \{d_n\},  \mathcal{P}_{\hat{i}}(\{d_n\})), 
\end{align}
$V$ continues until $d_{t+1}$ in $\mathcal{D}^*$ is identified. If no observation supports a $d_n$, the reasoning process will be stopped.

In our annotation, an observation $o$ is associated with only one $d$. However, our method employs an iterative reasoning pipeline. Initially, the perception module $W$ generates an explanation set $\hat{\mathcal{E}}_0$, linking all $\hat{o}$ to $d_0$. During the $t$-th iteration of $V$, the explanation set is $\hat{\mathcal{E}}_t$, where at least one $\hat{o}$ is linked to $d_t$. The final diagnosis explanation is the combination of ${\hat{\mathcal{E}}_0, \dots, \hat{\mathcal{E}}_T}$ and ${d_0, \dots, d_T}$, where $T$ represents the final iteration. In this combination, if an $\hat{o}$ is eventually processed in the iteration for $\hat{\mathcal{E}}_t$, the corresponding $(o, z, d)$ in all preceding ${\hat{\mathcal{E}}_0, \dots, \hat{\mathcal{E}}_{t-1}}$ will be removed. That is, $\hat{o}$ will always be possessed by the $d_t$ closest to the leaf PDD node.
\section{Experiments}

\subsection{Experimental Setup}\label{exp_setting}
We assess the reasoning capabilities of 7 recent LLMs from diverse families and model sizes, including 5 instruction-tuned models that are openly accessible: LLama3 8B and 70B \citep{llama3modelcard}, Zephyr 7B \citep{tunstall2023zephyr}, Mistral 7B \citep{jiang2023mistral}, and Mixtral 8$\times$7B \citep{jiang2023mistral}. We have also obtained access to private versions of the GPT-3.5 turbo \citep{GPT-turbo} and GPT-4 turbo \citep{GPT-4} \footnote{These two models are housed on a HIPPA-compliant instance within Microsoft Azure AI Studio. No data is transferred to either Microsoft or OpenAI. This secure environment enables us to safely conduct experiments with the MIMIC-IV dataset, in compliance with the Data Use Agreement.}, which are high-performance closed-source models. Each LLM is utilized to implement our baseline's narrowing-down, perception, and reasoning modules. The temperature is set to 0. For computing evaluation metrics, we use LLama3 8B with few-shot prompts to make correspondences between $\mathcal{O}$ and $\hat{\mathcal{O}}$ as well as to verify a match between predicted and ground-truth explanations (refer to the supplementary material for more details). 

\begin{table}[t]
\caption{Evaluation of diagnostic reasoning ability using $\mathcal{G}$ or $\mathcal{K}$ as input.}
\centering
\resizebox{1\columnwidth}{!}{
\begin{tabular}{llccccccc}
\toprule
&&\multicolumn{2}{c}{Diagnosis} &\multicolumn{3}{c}{Observation} & \multicolumn{2}{c}{Explanation}\\
\cmidrule(lr){3-4} \cmidrule(lr){5-7}  \cmidrule(lr){8-9} 
Task &Models & \textit{Acc}$^\text{cat}$ & \textit{Acc}$^\text{diag}$ & $\textit{Obs}^\text{pre}$ & $\textit{Obs}^\text{rec}$ & $\textit{Obs}^\text{comp}$ & $\textit{Exp}^\text{com}$ & $\textit{Exp}^\text{all}$ \\ 
\midrule
\multirow{7}{*}{With $\mathcal{G}$} &  Zephyr 7B &  0.274 & 0.151  & 0.123$_{\pm \text{0.200}}$ & 0.115$_{\pm \text{0.166}}$ & 0.092$_{\pm \text{0.108}}$ & 0.071$_{\pm \text{0.139}}$ & 0.014$_{\pm \text{0.037}}$ \\ 
&Mistral 7B &  0.507 & 0.306  & 0.211$_{\pm \text{0.190}}$ & 0.317$_{\pm \text{0.253}}$ & 0.173$_{\pm \text{0.157}}$ & 0.230$_{\pm \text{0.312}}$ & 0.062$_{\pm \text{0.088}}$ \\ 
&Mixtral 8$\times$7B  &  0.413 & 0.237  & 0.147$_{\pm \text{0.165}}$ & 0.266$_{\pm \text{0.261}}$ & 0.124$_{\pm \text{0.138}}$ & 0.144$_{\pm \text{0.268}}$ & 0.029$_{\pm \text{0.056}}$ \\ 
&LLama3 8B  &  0.569 & 0.364  & 0.248$_{\pm \text{0.157}}$ & 0.410$_{\pm \text{0.218}}$ & 0.211$_{\pm \text{0.138}}$ & 0.325$_{\pm \text{0.375}}$ & 0.087$_{\pm \text{0.118}}$ \\ 
&LLama3 70B  &  0.822 & 0.606  & 0.306$_{\pm \text{0.151}}$ & \textbf{0.543$_{\pm \text{0.183}}$} & 0.279$_{\pm \text{0.146}}$ & 0.409$_{\pm \text{0.328}}$ & 0.124$_{\pm \text{0.120}}$ \\ 
&GPT-3.5 turbo  &  0.679 & 0.455  & 0.389$_{\pm \text{0.212}}$ & 0.351$_{\pm \text{0.192}}$ & 0.275$_{\pm \text{0.167}}$ & 0.331$_{\pm \text{0.366}}$ & 0.103$_{\pm \text{0.127}}$ \\ 
&GPT-4 turbo  &  \textbf{0.804} & \textbf{0.610}  & \textbf{0.486$_{\pm \text{0.207}}$} & 0.481$_{\pm \text{0.180}}$ & \textbf{0.391$_{\pm \text{0.189}}$} & \textbf{0.481$_{\pm \text{0.362}}$} & \textbf{0.210$_{\pm \text{0.188}}$} \\
\midrule
\multirow{4}{*}{With $\mathcal{K}$} &  LLama3 8B &  0.576 & 0.344  & 0.235$_{\pm \text{0.162}}$ & 0.394$_{\pm \text{0.227}}$ & 0.199$_{\pm \text{0.142}}$ & 0.327$_{\pm \text{0.375}}$ & 0.087$_{\pm \text{0.114}}$ \\ 
& LLama3 70B &  0.786 & 0.652  & 0.268$_{\pm \text{0.147}}$ & \textbf{0.524$_{\pm \text{0.211}}$} & 0.258$_{\pm \text{0.142}}$ & 0.549$_{\pm \text{0.372}}$ & 0.152$_{\pm \text{0.130}}$ \\ 
& GPT-3.5 turbo &  0.652 & 0.413  & 0.347$_{\pm \text{0.241}}$ & 0.279$_{\pm \text{0.203}}$ & 0.232$_{\pm \text{0.184}}$ & 0.374$_{\pm \text{0.408}}$ & 0.121$_{\pm \text{0.152}}$ \\ 
& GPT-4 turbo  & \textbf{ 0.808} & \textbf{0.611}  & \textbf{0.470$_{\pm \text{0.209}}$} & 0.459$_{\pm \text{0.190}}$ & \textbf{0.371$_{\pm \text{0.192}}$} & \textbf{0.645$_{\pm \text{0.385}}$} & \textbf{0.273$_{\pm \text{0.216}}$} \\
\bottomrule
\end{tabular}}
\label{base_line}
\vspace*{-3mm}
\end{table}

\subsection{Results}
\textbf{Comparison among LLMs.} Table \ref{base_line} shows the performance of our baseline built on top of various LLMs. We first evaluate a variant of our task that takes graph $\mathcal{G} = \{\mathcal{G}_i\}$ consisting of only procedural flow as external knowledge instead of $\mathcal{K}$. Comparison between $\mathcal{G}$ and $\mathcal{K}$ demonstrates the importance of supplying premises with the model and LLMs' capability to make use of extensive external knowledge that may be superficially different from statements in $R$.  Subsequently, some models are evaluated with our task using $\mathcal{K}$. In addition to the metrics in Section \ref{sec:evaluation}, we also adopt the \textit{accuracy of disease category} $\textit{Acc}^{\text{cat}}$, which gives 1 when $\hat{i} = i$, as our baseline's performance depends on it.

With $\mathcal{G}$, we can see that GPT-4 achieves the best performance in most metrics, especially related to observations and explanations, surpassing LLama3 70B by a large margin. In terms of accuracy (in both category and diagnosis levels), LLama3 70B is comparable to GPT-4. LLama3 70B also has a higher $\textit{Obs}^\text{rec}$ but low $\textit{Obs}^\text{pre}$ and $\textit{Obs}^\text{comp}$, which means that this model tends to extract many observations. Models with high diagnostic accuracy are not necessarily excel in finding essential information in long text (i.e., observations) and generating reasons (i.e., explanations).

When $\mathcal{K}$ is given, all models show better diagnostic accuracy (in LLama3 70B) and explanations, while observations are slightly degraded (this may related to the instruction following ability due to the input length when giving $\mathcal{K}$ as input). GPT-4 with $\mathcal{K}$ enhances \textit{Acc}$^\text{diag}$, $\textit{Exp}^\text{com}$, and $\textit{Exp}^\text{all}$ scores. This suggests that premises and supporting edges are beneficial for diagnosis and explanation. Lower observational performance may indicate that the models lack the ability to associate premises and text in $R$, which are often superficially different though semantically consistent. 

\begin{table}[t]
\caption{Evaluation of diagnostic reasoning ability without external knowledge.}
\centering
\resizebox{0.98\columnwidth}{!}{
\begin{tabular}{llcccccc}
\toprule
& & &\multicolumn{3}{c}{Observation} & \multicolumn{2}{c}{Explanation}\\
 \cmidrule(lr){4-6}  \cmidrule(lr){7-8} 
Task & Models & \textit{Acc}$^\text{diag}$ & $\textit{Obs}^\text{pre}$ & $\textit{Obs}^\text{rec}$ & $\textit{Obs}^\text{comp}$ & $\textit{Exp}^\text{com}$ & $\textit{Exp}^\text{all}$ \\ 
\midrule
\multirow{4}{*}{With $\mathcal{D}^\star$} & LLama3 8B  & 0.070  & 0.154$_{\pm \text{0.139}}$ & 0.330$_{\pm \text{0.244}}$ & 0.135$_{\pm \text{0.122}}$ & 0.020$_{\pm \text{0.100}}$ & 0.004$_{\pm \text{0.016}}$ \\ 
& LLama3 70B  & 0.502  & 0.257$_{\pm \text{0.150}}$ & \textbf{0.509$_{\pm \text{0.213}}$} & 0.237$_{\pm \text{0.145}}$ & 0.138$_{\pm \text{0.209}}$ & 0.034$_{\pm \text{0.054}}$ \\ 
& GPT-3.5 turbo  & 0.223  & 0.164$_{\pm \text{0.242}}$ & 0.149$_{\pm \text{0.212}}$ & 0.116$_{\pm \text{0.174}}$  & 0.091$_{\pm \text{0.231}}$ & 0.025$_{\pm \text{0.065}}$\\ 
& GPT-4 turbo  & \textbf{0.636} & \textbf{0.461$_{\pm \text{0.206}}$} & 0.482$_{\pm \text{0.160}}$ & \textbf{0.378$_{\pm \text{0.174}}$} & \textbf{0.186$_{\pm \text{0.221}}$} & \textbf{0.074$_{\pm \text{0.090}}$} \\
\midrule
\multirow{4}{*}{No Knowledge} & LLama3 8B  & 0.023  & 0.137$_{\pm \text{0.159}}$ & 0.258$_{\pm \text{0.274}}$ & 0.119$_{\pm \text{0.141}}$ & 0.018$_{\pm \text{0.083}}$ & 0.006$_{\pm \text{0.026}}$ \\ 
& LLama3 70B  & 0.037  & 0.246 $_{\pm \text{0.148}}$ & \textbf{0.504$_{\pm \text{0.222}}$} & 0.227$_{\pm \text{0.148}}$ & 0.022$_{\pm \text{0.093}}$ & 0.007$_{\pm \text{0.030}}$ \\ 
& GPT-3.5 turbo  & 0.059  & 0.161$_{\pm \text{0.238}}$ & 0.148$_{\pm \text{0.215}}$ & 0.113$_{\pm \text{0.171}}$  & 0.036$_{\pm \text{0.131}}$ & 0.011$_{\pm \text{0.039}}$\\ 
& GPT-4 turbo  & \textbf{0.074} & \textbf{0.410$_{\pm \text{0.208}}$} & 0.443$_{\pm \text{0.191}}$ & \textbf{0.324$_{\pm \text{0.182}}$} & \textbf{0.047$_{\pm \text{0.143}}$} & \textbf{0.019$_{\pm \text{0.058}}$} \\
\bottomrule
\end{tabular}}
\label{table_naive}
\end{table}

LLMs may undergo inherent challenges for evaluation when no external knowledge is supplied. They may have the knowledge to diagnose but cannot make consistent observations and explanations that our task expects through $\mathcal{K}$. To explore this, we evaluate two settings: (1) giving $D^\star$ and (2) no knowledge is supplied to a model (shown in Table \ref{table_naive}). The prompts used for this setup are detailed in the supplementary material. We do not evaluate the accuracy of disease category prediction as it is basically the same as Table \ref{base_line}. We can clearly see that with $\mathcal{D}^\star$, GPT-4's diagnostic and observational scores are comparable to those of the task with $\mathcal{K}$, though explanatory performance is much worse. Without any external knowledge, the diagnostic accuracy is also inferior.\footnote{We understand this comparison is unfair, as the prompts differ. We intend to give a rough idea about the challenge without external knowledge.} The deteriorated performance can be attributed to inconsistent wording of diagnosis names, which makes evaluation tough. High observational scores imply that observations in $R$ can be identified without relying on external knowledge. There can be some cues to spot them. 

\begin{figure}[t]
\centering
\includegraphics[width=1\linewidth]{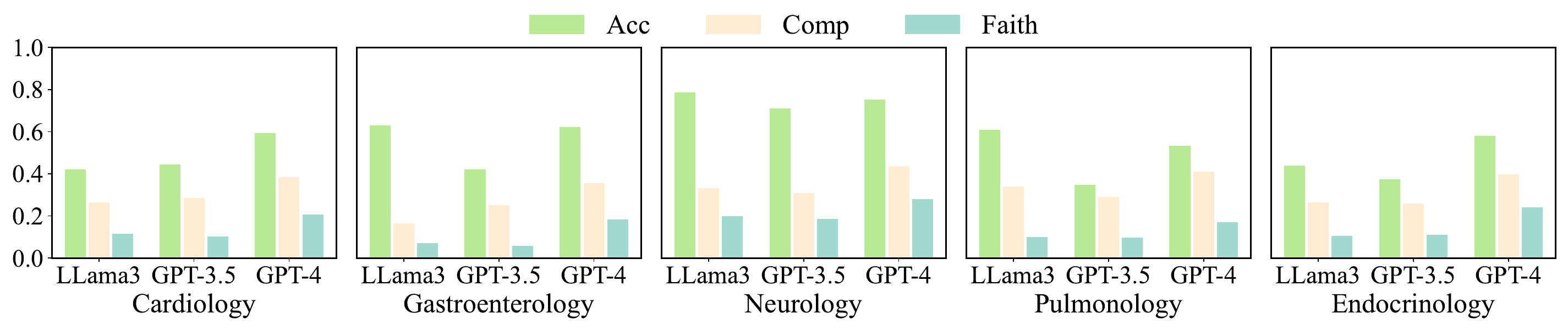}
\caption{Performance of LLama3 70B, GPT-3.5, and GPT-4 under different medical domains. We use the task with $\mathcal{G}$.}
\label{fig:each_disease}
\vspace*{-3mm}
\end{figure}

\textbf{Performance in individual domains.} Figure \ref{fig:each_disease} summarizes the performance of LLama3 70B, GPT-3.5, and GPT-4 across different medical domains, evaluated using \textit{Acc}$^\text{diag}$, $\textit{Obs}^\text{comp}$ (Comp), and $\textit{Exp}^\text{all}$ (Faith). Neurology gives the best diagnostic accuracy, where LLama3 achieved an accuracy of 0.779. GPT-4 also performed well (0.753). In terms of $\textit{Obs}^\text{comp}$ and $\textit{Exp}^\text{all}$, GPT-4's results were 0.437 and 0.280, respectively. However, GPT-4 yields a higher diagnostic accuracy score while a lower explanatory score, suggesting that the observations captured by the model or their rationalizations differ from human doctors.

\textbf{Diagnostic reasoning under conditions of incomplete observation.} In real-world scenarios, doctors often have to make diagnoses based on incomplete information. To explore this, we conducted experiments on the 73 amended cases which originally lack observation to the final diagnosis (refer to supplementary for detailed introduction of amended data point). One set of experiments used the unmodified original notes, labeled as "Original," while the other set used notes with added observations labeled as "Amended." We tested three models—Llama3 70B, GPT-3.5-turbo, and GPT-4 turbo—under two settings: one with only the procedural graph $\mathcal{G}$ and the other with the complete knowledge graph $\mathcal{K}$. The results are presented in Tables \ref{with_g} and \ref{with_k}. We can observe that in both $\mathcal{G}$ and $\mathcal{K}$ settings, the performance on the Amended data was consistently better across all metrics compared to the Original data. This suggests that even a single added observation can significantly impact the model’s diagnostic reasoning.

\begin{table}[t]
\caption{Amendment ablation study using $\mathcal{G}$.}
\centering
\resizebox{1\columnwidth}{!}{
\begin{tabular}{llccccccc}
\toprule
&&\multicolumn{2}{c}{Diagnosis} &\multicolumn{3}{c}{Observation} & \multicolumn{2}{c}{Explanation}\\
\cmidrule(lr){3-4} \cmidrule(lr){5-7}  \cmidrule(lr){8-9} 
Setting &Models & \textit{Acc}$^\text{cat}$ & \textit{Acc}$^\text{diag}$ & $\textit{Obs}^\text{pre}$ & $\textit{Obs}^\text{rec}$ & $\textit{Obs}^\text{comp}$ & $\textit{Exp}^\text{com}$ & $\textit{Exp}^\text{all}$ \\ 
\midrule
\multirow{3}{*}{Original} &  LLama3 70B &  0.547 & 0.273  & 0.225$_{\pm \text{0.143}}$ & 0.472$_{\pm \text{0.144}}$ & 0.253$_{\pm \text{0.138}}$ & 0.216$_{\pm \text{0.271}}$ & 0.073$_{\pm \text{0.087}}$ \\ 
&GPT-3.5 turbo &  0.507 & 0.273  & 0.393$_{\pm \text{0.216}}$ & 0.355$_{\pm \text{0.174}}$ & 0.278$_{\pm \text{0.151}}$ & 0.207$_{\pm \text{0.305}}$ & 0.062$_{\pm \text{0.093}}$ \\ 
&GPT-4 turbo  & 0.616 & 0.328 & 0.446$_{\pm \text{0.211}}$ & 0.418$_{\pm \text{0.164}}$ & 0.340$_{\pm \text{0.178}}$ & 0.242$_{\pm \text{0.324}}$ & 0.098$_{\pm \text{0.137}}$ \\ 
\midrule
\multirow{3}{*}{Amended} & LLama3 70B & 0.698 & 0.534  & 0.250$_{\pm \text{0.173}}$ & 0.507$_{\pm \text{0.134}}$ & 0.240$_{\pm \text{0.129}}$ & 0.296$_{\pm \text{0.354}}$ & 0.133$_{\pm \text{0.142}}$ \\ 
& GPT-3.5 turbo &  0.671 & 0.411  & 0.487$_{\pm \text{0.206}}$ & 0.351$_{\pm \text{0.152}}$ & 0.310$_{\pm \text{0.145}}$ & 0.272$_{\pm \text{0.321}}$ & 0.092$_{\pm \text{0.118}}$ \\ 
& GPT-4 turbo  & 0.726 & 0.547  & 0.546$_{\pm \text{0.184}}$ & 0.465$_{\pm \text{0.148}}$ & 0.412$_{\pm \text{0.171}}$ & 0.391$_{\pm \text{0.374}}$ & 0.180$_{\pm \text{0.186}}$ \\
\bottomrule
\end{tabular}}
\label{with_g}
\vspace*{-3mm}
\end{table}

\begin{table}[t]
\caption{Amendment ablation study using $\mathcal{K}$.}
\centering
\resizebox{1\columnwidth}{!}{
\begin{tabular}{llccccccc}
\toprule
&&\multicolumn{2}{c}{Diagnosis} &\multicolumn{3}{c}{Observation} & \multicolumn{2}{c}{Explanation}\\
\cmidrule(lr){3-4} \cmidrule(lr){5-7}  \cmidrule(lr){8-9} 
Setting &Models & \textit{Acc}$^\text{cat}$ & \textit{Acc}$^\text{diag}$ & $\textit{Obs}^\text{pre}$ & $\textit{Obs}^\text{rec}$ & $\textit{Obs}^\text{comp}$ & $\textit{Exp}^\text{com}$ & $\textit{Exp}^\text{all}$ \\ 
\midrule
\multirow{3}{*}{Original} &  LLama3 70B &  0.575 & 0.219  & 0.109$_{\pm \text{0.233}}$ & 0.443$_{\pm \text{0.171}}$ & 0.203$_{\pm \text{0.186}}$ & 0.304$_{\pm \text{0.388}}$ & 0.114$_{\pm \text{0.135}}$ \\ 
&GPT-3.5 turbo &  0.548 & 0.233  & 0.293$_{\pm \text{0.243}}$ & 0.218$_{\pm \text{0.198}}$ & 0.184$_{\pm \text{0.166}}$ & 0.251$_{\pm \text{0.357}}$ & 0.072$_{\pm \text{0.106}}$ \\ 
&GPT-4 turbo  & 0.616 & 0.260 & 0.452$_{\pm \text{0.241}}$ & 0.410$_{\pm \text{0.211}}$ & 0.349$_{\pm \text{0.223}}$ & 0.467$_{\pm \text{0.437}}$ & 0.220$_{\pm \text{0.256}}$ \\ 
\midrule
\multirow{3}{*}{Amended} & LLama3 70B &  0.685 & 0.537  & 0.261$_{\pm \text{0.195}}$ & 0.493$_{\pm \text{0.230}}$ & 0.277$_{\pm \text{0.171}}$ & 0.452$_{\pm \text{0.407}}$ & 0.185$_{\pm \text{0.194}}$ \\ 
& GPT-3.5 turbo &  0.657 & 0.465  & 0.390$_{\pm \text{0.227}}$ & 0.272$_{\pm \text{0.194}}$ & 0.232$_{\pm \text{0.156}}$ & 0.401$_{\pm \text{0.394}}$ & 0.127$_{\pm \text{0.145}}$ \\ 
&GPT-4 turbo  & 0.712 & 0.589 & 0.534$_{\pm \text{0.214}}$ & 0.452$_{\pm \text{0.180}}$ & 0.401$_{\pm \text{0.201}}$ & 0.607$_{\pm \text{0.442}}$ & 0.286$_{\pm \text{0.258}}$ \\ 
\bottomrule
\end{tabular}}
\label{with_k}
\vspace*{-3mm}
\end{table}

For Cardiology and Endocrinology, the diagnostic accuracy of the models is relatively low (GPT-4 achieved 0.458 and 0.468, respectively). Nevertheless, $\textit{Obs}^\text{comp}$ and $\textit{Exp}^\text{all}$ are relatively high. Endocrinology results in lower diagnostic accuracy and higher explanatory performance. A smaller gap may imply that in these two domains, successful predictions are associated with observations similar to those of human doctors, and the reasoning process may be analogous. Conversely, in Gastroenterology, higher \textit{Acc}$^\text{cat}$) is accompanied by lower $\textit{Obs}^\text{comp}$ and $\textit{Exp}^\text{all}$ (especially for LLama3), potentially indicating a significant divergence in the reasoning process from human doctors. Overall, DiReCT demonstrates that the degree of alignment between the model's diagnostic reasoning ability and that of human doctors varies across different medical domains.

\textbf{Reliability of automatic evaluation.} We randomly pick out 100 samples from DiReCT and their prediction by GPT-4 over the task with $\mathcal{G}$ to assess the consistency of our automated metrics to evaluate the observational and explanatory performance in Section \ref{sec:annotation} to human judgments. 
\begin{wraptable}{r}{0.60\textwidth}
\centering
\vspace*{-1mm}
\footnotesize
\caption{Consistency of automated evaluation with human judgments. Evaluated by mean and confidence interval (CI).}
\label{table_auto}
\resizebox{0.60\textwidth}{!}{%
\begin{tabular}{lcccc}
\toprule
&\multicolumn{2}{c}{Observation} & \multicolumn{2}{c}{Rationalization}\\
\cmidrule(lr){2-3} \cmidrule(lr){4-5} 
Model & Mean & 95\% CI & Mean & 95\% CI \\ 
\midrule
LLama3 8B &  0.887 & $0.844\sim0.878$  & 0.835 & $0.759\sim0.818$ \\ 
GPT-4 turbo &  0.902 & $0.830\sim0.863$  & 0.876 & $0.798\sim0.853$ \\ 
\bottomrule
\end{tabular}
}
\vspace*{-2mm}
\end{wraptable}
Three physicians joined this experiment. For each prediction $\hat{o} \in \hat{\mathcal{O}}$, they are asked to find a similar observation in ground truth $\mathcal{O}$. For explanatory metrics, they verify if each prediction $\hat{z} \in \hat{\mathcal{E}}$ for $\hat{o} \in \hat{\mathcal{O}}$ align with ground-truth $z \in \mathcal{E}$ corresponding to $o$. A prediction and a ground truth are deemed aligned for both assessments if at least two specialists agree. We compare LLama3's and GPT-4's judgments to explore if there is a gap between these LLMs. As summarized in Table \ref{table_auto}, GPT-4 achieves the best results, with LLama3 8B also displaying a similar performance. From these results, we argue that our automated evaluation metrics are consistent with human judgments, and LLama3 is sufficient for this evaluation, allowing the cost-efficient option. Detailed analysis is available in the supplementary material.

\begin{figure}[t]
\centering
\includegraphics[width=1\linewidth]{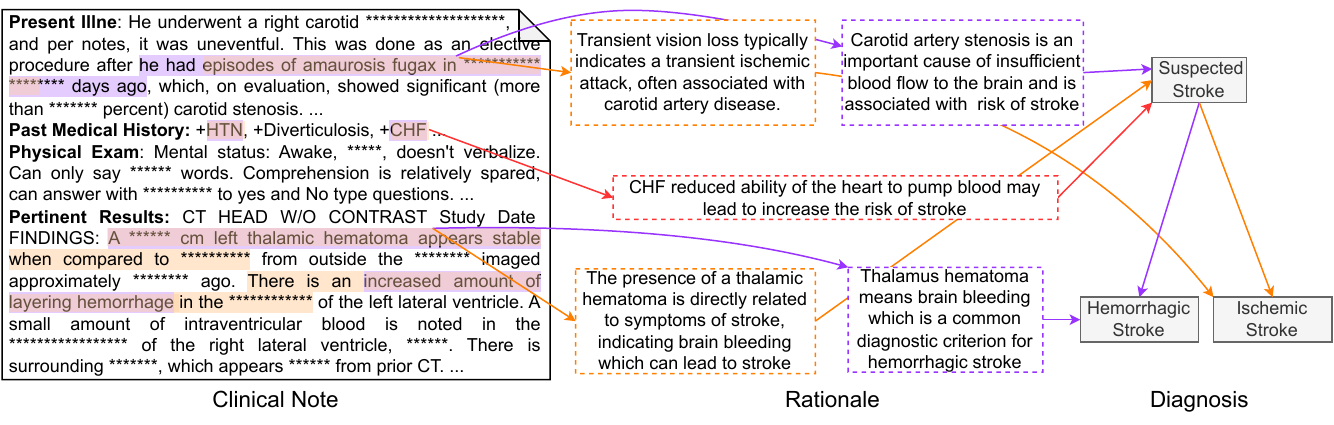}
\caption{An example prediction for a clinical note with PDD of \texttt{Hemorrhagic Stroke} by GPT-4.}
\label{fig:demo_stroke}
\vspace*{-2mm}
\end{figure}

\begin{figure}[t]
\centering
\includegraphics[width=1\textwidth]{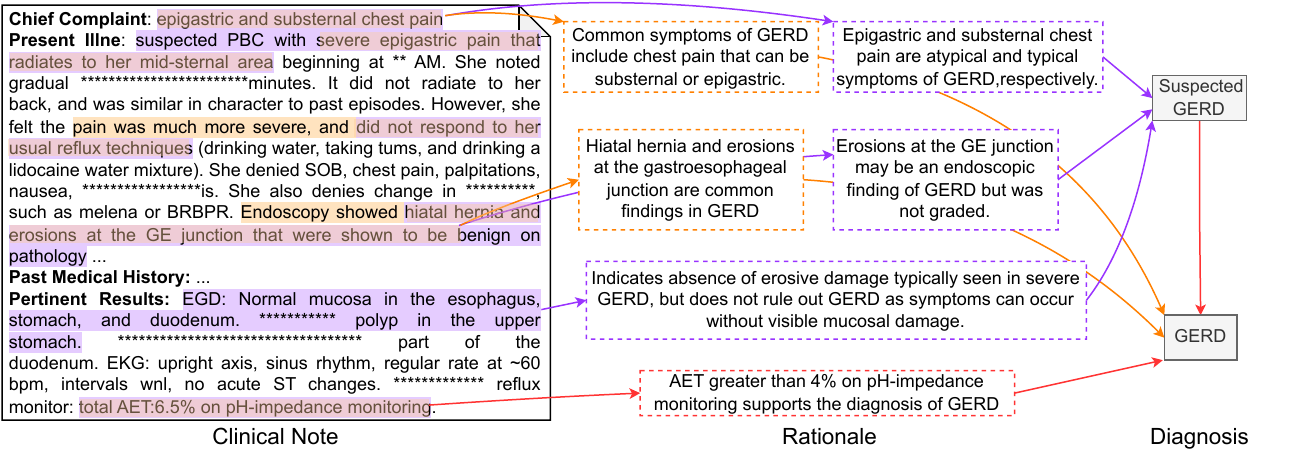}
\caption{An example prediction for a clinical note with PDD of \texttt{GERD} by GPT-4}
\label{fig:GERD}
\vspace*{-2mm}
\end{figure}

\textbf{Prediction examples.} Figure \ref{fig:demo_stroke} shows a sample generated by GPT-4. The ground-truth PDD of the input clinical note is \texttt{Hemorrhagic Stroke}. In this figure, purple, orange, and red indicate explanations only in the ground truth, only in prediction, and common in both, respectively; therefore, red is a successful prediction of an explanation, while purple and orange are a false negative and false positive. GPT-4 treats the observation of \texttt{aurosis fugax} as the criteria for diagnosing \texttt{Ischemic Stroke}. However, this observation only supports \texttt{Suspected Stroke}. Conversely, observation \texttt{thalamic hematoma}, which is the key indicator of \texttt{Hemorrhagic Stroke}, is regarded as a less important clue. Such observation-diagnosis correspondence errors lead to the model's misdiagnosis. In Figure \ref{fig:GERD}, we can observe that GPT-4 can find the key observation for the diagnosis of GERD, which is consistent with human in both observation and rationale. However, it still lacks the ability to identify all observations. More samples are available in the supplementary material.

\section{Conclusion and Limitations}\label{sec:conclusion}
We proposed DiReCT as the first benchmark for evaluating the diagnostic reasoning ability of LLMs with interpretability by supplying external knowledge as a graph. Our evaluations reveal a notable disparity between current leading-edge LLMs and human experts, underscoring the urgent need for AI models that can perform reliable and interpretable reasoning in clinical environments. DiReCT can be easily extended to more challenging settings by removing the knowledge graph from the input, facilitating evaluations of future LLMs. 

\textbf{Limitations.} DiReCT encompasses only a subset of disease categories and considers only one PDD, omitting the inter-diagnostic relationships due to their complexity---a significant challenge even for human doctors. Additionally, our baseline may not use optimal prompts or address issues related to hallucinations in task responses. Our dataset is solely intended for model evaluation but not for use in clinical environments. The use of the diagnostic knowledge graph is also limited to serving merely as a part of the input and once a knowledge graph is provided, the focus shifts to whether the LLM follows the graph's rules well (refer to supplementary). Future work will focus on constructing a more comprehensive disease dataset and developing an extensive diagnostic knowledge graph.


\begin{ack}
This work was supported by World Premier International Research Center Initiative (WPI), MEXT, Japan. This work is also supported by JST ACT-X Grant Number JPMJAX24C8, JSPS KAKENHI No. 24K20795 and No. JP23H00497, CREST Grant No. JPMJCR20D3, JST FOREST Grant No. JPMJFR216O, and Dalian Haichuang Project for Advanced Talents. 
\end{ack}

\newpage

\appendix
\vspace*{-23mm}
\section{Details of DiReCT}
\subsection{Data Statistics}

\begin{table}[ht]
\caption{Disease statistics of DiReCT.}
\centering
\resizebox{1\columnwidth}{!}{
\begin{tabular}{llcccc}
\toprule
Domains & Categories & \# samples& $|\mathcal{D}_i|$ & $|\mathcal{D}^{\star}_i|$ & References \\ 
\midrule
\multirow{7}{*}{Cardiology} &  Acute Coronary Syndromes &  65 & 6  & 3 &  \citep{byrne20242023,kitaoka2020jcs} \\ 
&  Aortic Dissection &  14 & 3  & 2 & \citep{writing20222022}  \\ 
&  Atrial Fibrillation &  10 & 3  & 2 & \citep{joglar20242023}  \\ 
&  Cardiomyopathy &  9 & 5  & 4 & \citep{ommen20202020}  \\ 
&  Heart Failure &  52 & 6  & 3 & \citep{heidenreich20222022}  \\ 
&  Hyperlipidemia &  2 & 2  & 1 &  \citep{su2021current,mach20202019} \\ 
&  Hypertension &  32 & 2  & 1 & \citep{unger20202020} \\ 
\midrule
\multirow{4}{*}{Gastroenterology} &  Gastritis &  27 & 5  & 3 & \citep{shah2021aga,ChineseGastritisGuidelines2023,banks2019british,chow2010acute} \\ 
&  Gastroesophageal Reflux Disease &  41 & 2  & 1 & \citep{gyawali2024updates} \\ 
&  Peptic Ulcer Disease &  28 & 3  & 2 & \citep{kavitt2019diagnosis,tarasconi2020perforated} \\ 
&  Upper Gastrointestinal Bleeding &  7 & 2 & 1 & \citep{barkun2019management}  \\ 
\midrule
\multirow{5}{*}{Neurology} &  Alzheimer &  10 & 2  & 1 & \citep{mckhann1984clinical}  \\ 
&  Epilepsy &  8 & 3  & 2 & \citep{Epilepsy2018}  \\ 
&  Migraine &  4 & 3  & 2 & \citep{lipton2001migraine,Eigenbrodt2021,IHS2018}  \\ 
&  Multiple Sclerosis & 27 & 6  & 4 & \citep{lublin2005clinical,Brownlee2017}  \\ 
& Stroke &  28 & 3  & 2 & \citep{kleindorfer20212021}  \\ 
\midrule
\multirow{5}{*}{Pulmonology} &  Asthma &  13 & 7  & 5 & \citep{qaseem2011diagnosis,bateman2007global,baos2018nonallergic} \\ 
&  COPD &  19 & 6  & 4 & \citep{gupta2013guidelines}  \\ 
&  Pneumonia &  20 & 4  & 2 & \citep{olson2020diagnosis,recommendations2012guidelines,niederman2001guidelines} \\ 
&  Pulmonary Embolism &  35 & 5  & 3 & \citep{konstantinides20202019} \\ 
& Tuberculosis &  5 & 3  & 2 & \citep{lewinsohn2017official}  \\ 
\midrule
\multirow{4}{*}{Endocrinology} & Adrenal Insufficiency &  20 & 4  & 3 & \citep{charmandari2014adrenal,yanase2016diagnosis,bornstein2016diagnosis} \\ 
&  Diabetes &  13 & 4  & 2 & \citep{elsayed20234}  \\ 
&  Pituitary &  12 & 4 & 3 & \citep{tritos2023diagnosis,drummond2019clinical,cooper2012subclinical,mayson2014silent}  \\ 
&  Thyroid Disease &  10 &  6 & 4 & \citep{alexandererik20172017} \\ 
\bottomrule
\end{tabular}}
\label{statistic}
\vspace*{-3mm}
\end{table}

Table \ref{statistic} provides a detailed breakdown of the disease categories included in DiReCT. The column labeled \# samples indicates the number of data points. The symbols $|\mathcal{D}_i|$ and $|\mathcal{D}^{\star}_i|$ denote the total number of diagnoses (diseases) and PDDs, respectively. Existing guidelines for diagnosing diseases were used as References, forming the foundation for constructing the diagnostic knowledge graphs. As some premise may not included in the referred guidelines. During annotation, physicians will incorporate their own knowledge to complete the knowledge graph. 

\begin{table}[t]
\centering
\begin{tabular}{|>{\centering\arraybackslash}p{1.5cm}|p{5cm}|>{\centering\arraybackslash}p{1.5cm}|p{5cm}|}
\hline
\textbf{Notation} & \textbf{Description}&\textbf{Notation}& \textbf{Description}\\ \hline
$R$ & The whole content of input note. & $U$ & The narrowing-down module.\\ \hline
$r$ & One data section of the input note. & $W$ & The perception module. \\ \hline
$\mathcal{D}$ & Disease collection.& $V$ & The reasoning module. \\ \hline
$\mathcal{D}^{\star}$ & PDD collection. & $\{d_n\}$& Collection of children diagnosis. \\ \hline
$\mathcal{D}^{\star}_i$ & PDD collection for disease $i$. & $Acc^{diag}$ & Diagnosis accuracy for $d^{\star}$.\\ \hline
$d^{\star}$ & A PDD disease. &$Acc^{cat}$& Diagnosis accuracy for category.\\ \hline
$d_t$ & Diagnosis at $t$-th iteration. & $Obs^{pre}$& Precision of observation.\\ \hline
$d$ & A diagnosis in $\mathcal{G}$. & $Obs^{rec}$& Recall of observation.\\ \hline
$\mathcal{G}$ &  Procedural graphs.& $Obs^{comp}$& Completeness of observation.\\ \hline
$g_i$ & Procedural subgraph for disease & $Exp^{com}$& Completeness of explanation.\\ \hline
$\mathcal{K}$ &  knowledge graphs.& $Exp^{all}$& Completeness of all explanation.\\ \hline
$k_i$ & knowledge subgraphs for disease $i$. & $M$& An language model.\\ \hline
$\mathcal{P}_i$ & Supporting edge collection for $k_i$.&$\mathcal{E}$ & Collection of annotated deductions.\\ \hline
$p$ & A premise defined in $K$.&& \\ \hline
$\mathcal{F}_i$ & Procedural edge collection for $g_i$.&& \\ \hline
$\mathcal{O}$ & Collection of annotated observation.&& \\ \hline
$o$ & An annotated observation.&& \\ \hline
$z$ & Rationale for a deduction.&& \\ \hline
$d_0$ & Root diagnosis for $g_i$.&& \\ \hline

\end{tabular}
\caption{Notations defined in this paper.}
\label{tab:notations}
\end{table}

\subsection{Amended Data Points}
Our proposed dataset aims to evaluate whether LLMs can provide a complete diagnostic reasoning process comparable to that of human doctors. To achieve this, we intended to select notes from the MIMIC database that contain comprehensive signs and symptoms as observations, enabling physicians to annotate the notes leading to a final PDD. For disease category like heart failure, MIMIC offers ample data, allowing us to choose notes with complete observations. However, for PDDs such as bacterial pneumonia, the number of relevant notes is limited, and many lack critical evidence necessary for diagnosis (e.g., sputum culture). We observed that in some notes, the section under the title 'sputum culture' was left blank. We suspect that this might be due to some information being missed in MIMIC. To annotate such cases, we ask physicians add the necessary observations to support the diagnosis. In total, we made amendments to 73 notes. These notes all lacked evidence for a final PDD diagnosis, and in each note, only one observation was added as evidence. Thus, the modifications to the original content of the notes were minimal. For example, in a note where the PDD is bacterial pneumonia, we only added the following description under 'pertinent results': 'Multiple organisms consistent with Haemophilus influenzae.'

To better illustrate the structure of our dataset and identify which data has been amended and what content has been added, we have provided a detailed CSV file on GitHub (https://github.com/wbw520/DiReCT/tree/master/utils/data\_loading\_analysisi). This file contains six columns, which record the following information: Disease Category, PDD, Data Root, Whether Amended, Amended Part, and Amended Content. The Data Root column records the path and filename of each note. We have stored the original note information and our annotations within a JSON file. The version submitted for review to PhysioNet follows the same storage format. In the Whether Amended column, notes that have been amended are marked as 'Yes,' with the Amended Part and Amended Content columns specifying which part of the note was modified and what content was added. Additionally, we have provided several synthetic annotated samples (non-MIMIC data) on GitHub, along with detailed instructions on the format of the annotated data and how to parse each JSON file.

\subsection{Structure of Knowledge Graph}
\begin{figure}[t]
\centering
\includegraphics[width=1\textwidth]{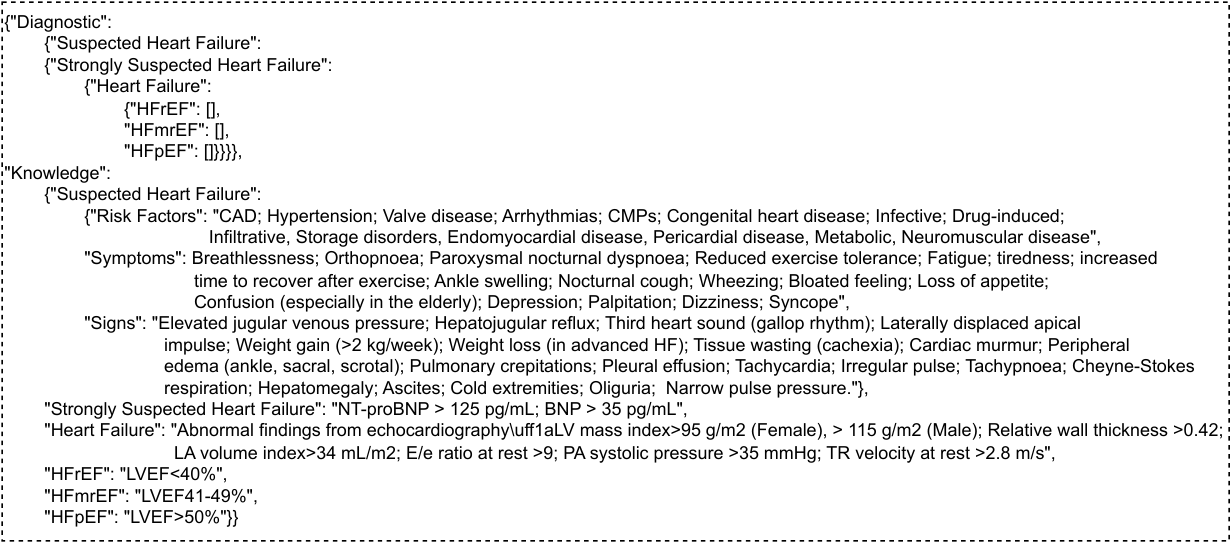}
\caption{A sample of knowledge graph for \texttt{Heart Failure}. Each premise under the key of "Knowledge" is separated with ";".}
\label{fig:KG}
\end{figure}

We first show the notations definition on Table \ref{tab:notations}. The entire knowledge graph, denoted as $\mathcal{K}$, is stored in separate JSON files, each corresponding to a specific disease category $i$ as $\mathcal{K}_i$. Each $\mathcal{K}_i$ comprises a procedural graph $\mathcal{G}_i$ and the corresponding premise $p$ for each disease. As illustrated in Figure \ref{fig:KG}, the procedural graph $\mathcal{G}_i$ is stored under the key "Diagnostic" in a dictionary structure. A key with an empty list as its value indicates a leaf diagnostic node as $d^{\star}$. The premise for each disease is saved under the key of "Knowledge" with the corresponding disease name as an index. For all the root nodes (e.g., \texttt{Suspected Heart Failure}), we further divide the premise into "Risk Factors", "Symptoms", and "Signs". Note that each premise is separated by ";".

Our knowledge graph was directly constructed by human physicians who followed authoritative diagnostic guidelines and incorporated their clinical experience. For Cardiology, Gastroenterology, Neurology, Pulmonology, and Endocrinology, the knowledge graph was built by 2, 1, 2, 2, and 1 specialists from the respective departments. The construction process involved first defining the procedural graph $g_i$ for each category, followed by supplementing $g_i$ with the detailed premises corresponding to each diagnosis d to build $k_i$. The complete knowledge graphs are available on GitHub (https://github.com/wbw520/DiReCT/tree/master/utils/data\_annotation).

\subsection{Annotation and Tools}
\begin{figure}[t]
\centering
\includegraphics[width=1\textwidth]{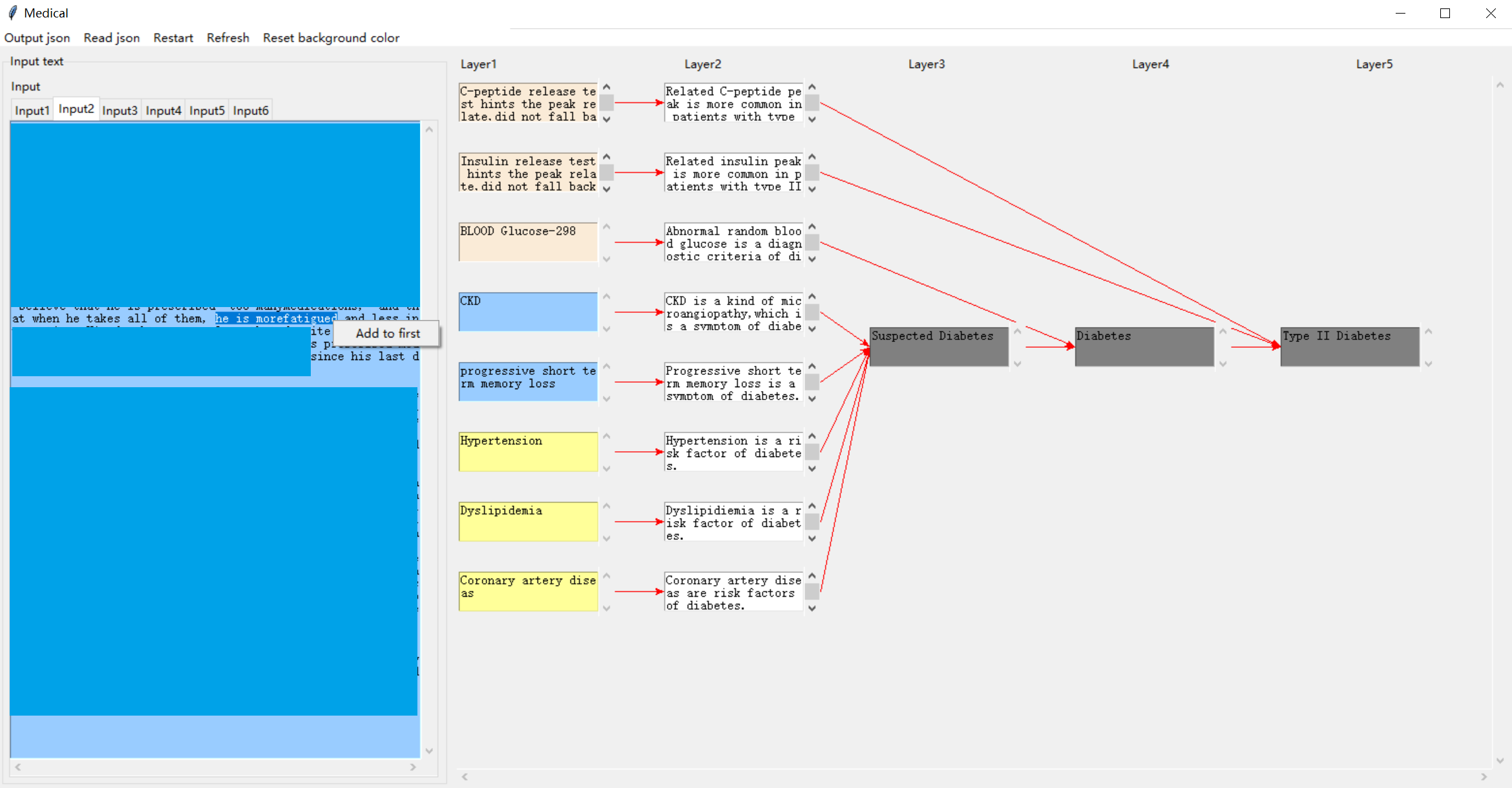}
\caption{Demonstration of our annotation tool.}
\label{fig:annotation}
\end{figure}

We have developed proprietary software for annotation purposes. As depicted in Figure \ref{fig:annotation}, annotators are presented with the original text as observations $o$ and are required to provide rationales ($z$) to explain why a particular observation $o$ supports a disease $d$. The left section of the figure, labeled Input1 to Input6, corresponds to different parts of the clinical note, specifically the chief complaint, history of present illness, past medical history, family history, physical exam, and pertinent results, respectively. Annotators will add the raw text into the first layer by left-clicking and dragging to select the original text, then right-clicking to add it. After each observation, a white box will be used to record the rationales. Finally, a connection will be made from each rationale to a disease, represented in a grey box. The annotation process strictly follow the knowledge graph. Both the final annotation and the raw clinical note will be saved in a JSON file. We provide the code to compile these annotations and detailed instructions for using our tool on GitHub.

\subsection{Access to DiReCT}
Implementation code and annotation tool are available through https://github.com/wbw520/DiReCT. Data will be released through PhysioNet due to safety issues according to the license of MIMIC-IV (PhysioNet Credentialed Health Data License 1.5.0). We will use the same license for DiReCT. The download link will be accessible via GitHub. We confirm that this GitHub link and data link are always accessible. We confirm that we will bear all responsibility in case of violation of rights.

\section{Implementation of Baseline Method}
\subsection{Prompt Settings}

\begin{table}[t]
\caption{Prompt for \textit{narrowing-down} module.}
\centering
\resizebox{1\columnwidth}{!}{
\begin{tabular}{l}
\toprule
Input Prompt \\
\midrule
Suppose you are one of the greatest AI scientists and medical expert. Let us think step by step.\\ 
You will review a clinical 'Note' and your 'Response' is to diagnose the disease that the patient have for this admission. \\
All possible disease options are in a list structure: \color{red}{\{disease\_option\}}. \\
Note that you can only choose one disease from the disease options and directly output the origin name of that disease. \\
Now, start to complete your task.\\
Don't output any information other than your 'Response'.\\
'Note':\\
\color{red}{\{note\}}\\
Your 'Response': \\
\bottomrule
\end{tabular}
}
\label{narrowing}
\vspace*{-5mm}
\end{table}

\begin{table}[t]
\caption{Prompt for \textit{perception} module.}
\centering
\resizebox{1\columnwidth}{!}{
\begin{tabular}{l}
\toprule
Input Prompt \\
\midrule
Suppose you are one of the greatest AI scientists and medical expert. Let us think step by step.\\ 
You will review a part of clinical "Note" from a patient. \\
The disease for which the patient was admitted to hospital this time is \color{red}{\{disease\}}.\\
Your task is to extract the original text as confidence "Observations" that lead to \color{red}{\{disease\}}.\\
\color{blue}{Here are some premise for the diagnosis of this disease category. You can refer them for your task. Premise are: \{premise\}} \\
Note that you also need to briefly provide the "Reason" for your extraction.\\
Note that both "Observations" and "Reason" should be string.\\
Note that your "Response" should be a list structure as following \\
: [["Observation", "Reason"], ......, ["Observation", "Reason"]]\\
Note that if you can't find any "Observation" your "Response" should be: [].\\
Now, start to complete your task. \\
Note that you should not output any information other than your "Response".\\
"Note":\\
\color{red}{\{note\}}\\
Note that you should not output any information other than your "Response".\\
Your "Response":\\
\bottomrule
\end{tabular}
}
\label{perception}
\vspace*{-5mm}
\end{table}

\begin{table}[t]
\caption{Prompt for \textit{reasoning} module.}
\centering
\resizebox{1\columnwidth}{!}{
\begin{tabular}{l}
\toprule
Input Prompt \\
\midrule
Suppose you are one of the greatest AI scientists and medical expert. Let us think step by step.\\ 
You will receive a list of "Observations" from a clinical "Note". These "Observations" are possible support to diagnose \color{red}{\{disease\}}.\\
Based on these "Observations", you need to diagnose the "Disease" from the following options: \color{red}{\{disease\_option\}}. \\
\color{blue}{Here are some golden standards to discriminate diseases. You can refer them for your task. Golden standards are: \{premise\}} \\
Note that you can only choose one "Disease" from the disease options and directly output the name in disease options.\\
Note that you also required to select the "Observations" that satisfy the golden standard to diagnose the "Disease" you choose.\\
Note that you also required to provide the "Reason" for your choice.\\
Note that your "Response" should be a list structure as following\\
:[["Observation", "Reason", "Disease"], ......, ["Observation", "Reason", "Disease"]]\\
Note that if you can't find any "Observation" to support a disease option, your "Response" should be: None\\
Now, start to complete your task. \\
Note that you should not output any information other than your "Response".\\
"Observations":\\
\color{red}{\{observation\}}\\
Note that you should not output any information other than your "Response".\\
Your "Response":\\
\bottomrule
\end{tabular}
}
\label{reasoning}
\vspace*{-2mm}
\end{table}

In this section, we demonstrate the prompt we used for each module (From Table \ref{narrowing}-\ref{reasoning} for \textit{narrowing-down}, \textit{perception}, and \textit{reasoning} module, respectively). 

In Table \ref{narrowing}, \{disease\_option\} is the name for all disease categories, and \{note\} is the content for the whole clinical note. The response for the model is the name of a possible disease $\hat{i}$.

In Table \ref{perception}, \{disease\} is the disease category name predicted in \textit{narrowing-down}. The content marked blue is the premise, which is only provided during the $\mathcal{K}$ setting. In this module, \{premise\} is offered with all information in the knowledge graph. Different to \textit{narrowing-down}, \{note\} is implemented for each clinical data $R=\{r\}$ and the outputs are combined together for $\hat{\mathcal{O}}$ and $\hat{\mathcal{E}}$.

In Table \ref{reasoning},  \{disease\} is the disease category name and \{disease\_option\} is consisted by the children nodes $\{d_n\}$. Similarly, the premise on the blue is only available for the $\mathcal{K}$ setting. It provides the premise that are criteria for the diagnosis of each children node. \{observation\} is the extracted $\hat{\mathcal{O}}$ in previous step. We provide all the prompts and the complete implementation code on GitHub.

\subsection{Diagnostic Reasoning Under Conditions of Incomplete Observation}
In real-world scenarios, doctors often have to make diagnoses based on incomplete information. To explore this, we conducted experiments on the 73 amended cases. One set of experiments used the unmodified original notes, labeled as "Original," while the other set used notes with added observations, labeled as "Amended." We tested three models—Llama3 70B, GPT-3.5-turbo, and GPT-4 turbo—under two settings: one with only the procedural graph $\mathcal{G}$ and the other with the complete knowledge graph $\mathcal{K}$. The results are presented in Tables \ref{with_g} and \ref{with_k}. We can observe that in both $\mathcal{G}$ and $\mathcal{K}$ settings, the performance on the Amended data was consistently better across all metrics compared to the Original data. This suggests that even a single added observation can significantly impact the model’s diagnostic reasoning.

Additionally, we found that under the Amended data, using $\mathcal{K}$ led to both better diagnostic outcomes and improved explanability, aligning with the analysis in our paper. However, when using $\mathcal{K}$ on the Original data, while explanability improved, diagnostic accuracy actually decreased.

We conducted a detailed analysis of the 73 Original data's results from GPT-4. We found that GPT-4 was still able to correctly deduce the final PDD in 24 cases using $\mathcal{G}$ and 19 cases using $\mathcal{K}$. This indicates that the model possesses some level of uncertain reasoning capability. However, upon further inspection, we found that in some cases, the model used completely irrational observations as evidence, such as directly using "cough" as evidence for diagnosing "bacterial pneumonia". Additionally, there were 7 cases using $\mathcal{G}$ and 13 cases using $\mathcal{K}$ where the reasoning stopped before the final PDD diagnosis. This suggests that the model recognized the lack of sufficient evidence to derive the PDD and adhered faithfully to the diagnostic knowledge graph. Moreover, using appeared to help the model better understand this limitation, however, decrease the accuracy. These results indicate that employing the knowledge graph acts more like a trade-off: using only $\mathcal{G}$ results in a higher tendency for uncertain reasoning, while using the full $\mathcal{K}$ makes the model more cautious.

\textbf{Limitation of current implementation.} Once a knowledge graph is provided, the focus shifts to whether the LLM follows the graph's rules well. However, we consider the knowledge graph as an inferential framework rather than a set of rules. This framework provides decision-making paths for the LLM, but the LLM still needs to perform reasoning within it. Even when strictly following the knowledge graph, the LLM still needs to perform semantic analysis and context understanding in order to select the node that best suits the current situation among multiple possible paths (sub-nodes) in the knowledge graph. Therefore, the role of the LLM in this process is not merely to 'follow the rules,' but to make logical path selections based on its understanding of the input data, which itself is a reflection of reasoning ability. This often requires the model or algorithm to consider previous steps in each stage of reasoning and to update observations accordingly. Even revise or backtrack the diagnosis step. However, our baseline method did not account for this and thus cannot fully exploit this capability of the LLM, which is a current limitation. We did try some designs to give those abilities to LLMs, such as providing previous steps of reasoning for the current stage as input prompts or update observations. However, even GPT-4 cannot show high instruction following ability to realize them (maybe the input is too long or prompt setting problems).

For evaluation, we used diagnostic processes annotated by human doctors as ground truth. Therefore, whether the KG is provided or not, the model's output needs to align with the ground truth. Our dataset allows for evaluation with and without the KG, but our baseline method is not effective at handling scenarios without the KG (this is much more challenging). How to utilize and explore the LLM's reasoning ability in this scenario is one of our future research directions.

\subsection{Details of Automatic Evaluation}
The automatic evaluation is realized by LLama3 8B. We demonstrate the prompt for this implement in Table \ref{observation_eval} (for observation) and Table \ref{rationale_eval} (for rationalization). Note that we do not use few-shot samples for the evaluation of observation. In Table \ref{observation_eval}, \{gt\_observation\} and \{pred\_observation\} are from model prediction and ground-truth. As this is a simple similarity comparison task to discriminate whether the model finds similar observations to humans, LLama3 itself have such ability. We do not strict to exactly match due to the difference in length of extracted raw text (as long as the observation expresses the same description). In Table \ref{rationale_eval}, \{gt\_reasoning\} and \{pred\_reasoning\} are from model prediction and ground-truth. We require the rationale to be complete (content of the expression can be understood from the rationale alone) and meaningful; therefore, we provide five samples for this evaluation. We also provide all the prompts and the complete implementation code on GitHub.

\begin{table}[t]
\caption{Prompt for evaluation of observation.}
\centering
\resizebox{0.75\columnwidth}{!}{
\begin{tabular}{l}
\toprule
Input Prompt \\
\midrule
Suppose you are one of the greatest AI scientists and medical expert. Let us think step by step.\\
You will receive two "Observations" extracted from a patient's clinical note. \\
Your task is to discriminate whether they textually description is similar? \\
Note that "Response" should be one selection from "Yes" or "No".\\
Now, start to complete your task.\\
Don't output any information other than your "Response".\\
"Observation 1": \color{red}{\{gt\_observation\}}\\
"Observation 2": \color{red}{\{pred\_observation\}}\\
Your "Response":\\
\bottomrule
\end{tabular}
}
\label{observation_eval}
\end{table}

\begin{table}[t]
\caption{Prompt for evaluation of rationalization.}
\centering
\resizebox{1\columnwidth}{!}{
\begin{tabular}{l}
\toprule
Input Prompt \\
\midrule
Suppose you are one of the greatest AI scientists and medical expert. Let us think step by step.\\
You will receive two "Reasoning" for the explanation of why an observation cause a disease. \\
Your task is to discriminate whether they explain a similar medical diagnosis premise? \\
Note that "Response" should be one selection from "Yes" or "No".\\
Here are some samples: \\
Sample 1:\\
    "Reasoning 1": Facial sagging is a classic symptom of stroke\\
    "Reasoning 2": Indicates possible facial nerve palsy, a common symptom of stroke\\
    "Response": Yes\\
Sample 2: \\
    "Reasoning 1": Family history of Diabetes is an important factor\\
    "Reasoning 2": Patient's mother had a history of Diabetes, indicating a possible genetic predisposition to stroke\\
    "Response": Yes\\
Sample 3:\\
    "Reasoning 1": headache is one of the common symptoms of HTN\\
    "Reasoning 2": Possible symptom of HTN\\
    "Response": No\\
Sample 4:\\
    "Reasoning 1": Acute bleeding is one of the typical symptoms of hemorrhagic stroke\\
    "Reasoning 2": The presence of high-density areas on Non-contrast CT Scan is a golden standard for Hemorrhagic Stroke\\
    "Response": No\\
Sample 5:\\
    "Reasoning 1": Loss of strength on one side of the body, especially when compared to the other side, is a common sign of stroke\\
    "Reasoning 2": Supports ischemic stroke diagnosis\\
    "Response": No\\
Now, start to complete your task.\\
Don't output any information other than your "Response".\\
"Reasoning 1": \color{red}{\{gt\_reasoning\}}\\
"Reasoning 2": \color{red}{\{pred\_reasoning\}}\\
Your "Response":\\
\bottomrule
\end{tabular}
}
\label{rationale_eval}
\end{table}

For human evaluation, among the three specialists, two are from Cardiology and one is from Gastroenterology. Given that the notes originate from different medical domains, there is a possibility that the specialists may not be entirely accurate. However, this evaluation does not demand highly specialized knowledge, and it can be adequately covered by their expertise.

We also included an experimental result comparing the judgment differences between Llama3 8B and GPT-4 Turbo. The evaluation was performed on the diagnostic outcomes (across the entire dataset) from Llama3 70B and GPT-4 Turbo, using $\mathcal{G}$ as additional knowledge. We calculated the consistency rate for matching observations and the corresponding rationalization. As shown in Table \ref{human}, the differences in judgment between the two models are not obvious and are more consistent in observation discrimination. There are also some variations across different disease domains, with the highest similarity in observation discrimination found in Endocrinology, while the rationalization is most similar in Neurology.

Additionally, we provided results using GPT-4 Turbo for automatic evaluation, compared to those shown in Table \ref{difference} (which used Llama3 8B). The results indicate that GPT-4 Turbo tends to yield higher observation matching and more stringent rationalization discrimination. However, the largest difference does not exceed 5\%. Considering the cost of GPT-4, Llama3 8B is a more efficient option.

\begin{table}[t]
\caption{Judgement consistency between LLama3 8B and GPT-4 turbo.}
\centering
\resizebox{0.78\columnwidth}{!}{
\begin{tabular}{lcccc}
\toprule
&\multicolumn{2}{c}{LLama3 70B} & \multicolumn{2}{c}{GPT-4 turbo}\\
\cmidrule(lr){2-3} \cmidrule(lr){4-5} 
Domain & Observation & Rationalization & Observation & Rationalization \\ 
\midrule
Cardiology &  0.885$_{\pm \text{0.095}}$ & 0.761$_{\pm \text{0.268}}$  & 0.827$_{\pm \text{0.146}}$ & 0.861$_{\pm \text{0.273}}$  \\ 
Gastroenterology & 0.862$_{\pm \text{0.088}}$ & 0.676$_{\pm \text{0.361}}$  & 0.810$_{\pm \text{0.167}}$ & 0.755$_{\pm \text{0.316}}$ \\ 
Neurology &  0.846$_{\pm \text{0.090}}$ & 0.831$_{\pm \text{0.211}}$  & 0.856$_{\pm \text{0.124}}$ & 0.963$_{\pm \text{0.106}}$  \\ 
Pulmonology &  0.808$_{\pm \text{0.131}}$ & 0.703$_{\pm \text{0.317}}$  & 0.786$_{\pm \text{0.152}}$ & 0.779$_{\pm \text{0.287}}$  \\ 
Endocrinology &  0.911$_{\pm \text{0.104}}$ & 0.783$_{\pm \text{0.304}}$  & 0.868$_{\pm \text{0.145}}$ & 0.793$_{\pm \text{0.340}}$  \\ 
Overall & 0.869$_{\pm \text{0.102}}$ & 0.734$_{\pm \text{0.321}}$  & 0.838$_{\pm \text{0.144}}$ & 0.806$_{\pm \text{0.305}}$  \\ 
\bottomrule
\end{tabular}}
\label{human}
\vspace*{-3mm}
\end{table}

\begin{table}[t]
\caption{Result of using GPT-4 turbo and LLama3 8B for automatic evaluation.}
\centering
\resizebox{0.92\columnwidth}{!}{
\begin{tabular}{llccccc}
\toprule
& &\multicolumn{3}{c}{Observation} & \multicolumn{2}{c}{Explanation}\\
\cmidrule(lr){3-5}  \cmidrule(lr){6-7} 
Judgement &Models & $\textit{Obs}^\text{pre}$ & $\textit{Obs}^\text{rec}$ & $\textit{Obs}^\text{comp}$ & $\textit{Exp}^\text{com}$ & $\textit{Exp}^\text{all}$ \\ 
\midrule
\multirow{2}{*}{GPT-4 turbo} &  LLama3 70B  & 0.317$_{\pm \text{0.161}}$ & 0.576$_{\pm \text{0.195}}$ & 0.294$_{\pm \text{0.159}}$ & 0.348$_{\pm \text{0.300}}$ & 0.107$_{\pm \text{0.118}}$ \\ 
&GPT-4 turbo & 0.465$_{\pm \text{0.190}}$ & 0.514$_{\pm \text{0.157}}$ & 0.408$_{\pm \text{0.201}}$ & 0.437$_{\pm \text{0.335}}$ & 0.187$_{\pm \text{0.191}}$ \\ 
\midrule
\multirow{2}{*}{LLama3 8B} & LLama3 70B  & 0.277$_{\pm \text{0.146}}$ & 0.537$_{\pm \text{0.192}}$ & 0.256$_{\pm \text{0.142}}$ & 0.395$_{\pm \text{0.320}}$ & 0.112$_{\pm \text{0.110}}$ \\ 
& GPT-4 turbo  & 0.446$_{\pm \text{0.207}}$ & 0.491$_{\pm \text{0.180}}$ & 0.371$_{\pm \text{0.186}}$ & 0.475$_{\pm \text{0.363}}$ & 0.199$_{\pm \text{0.181}}$ \\ 
\bottomrule
\end{tabular}}
\label{difference}
\vspace*{-3mm}
\end{table}

\begin{figure}[t]
\centering
\includegraphics[width=1\textwidth]{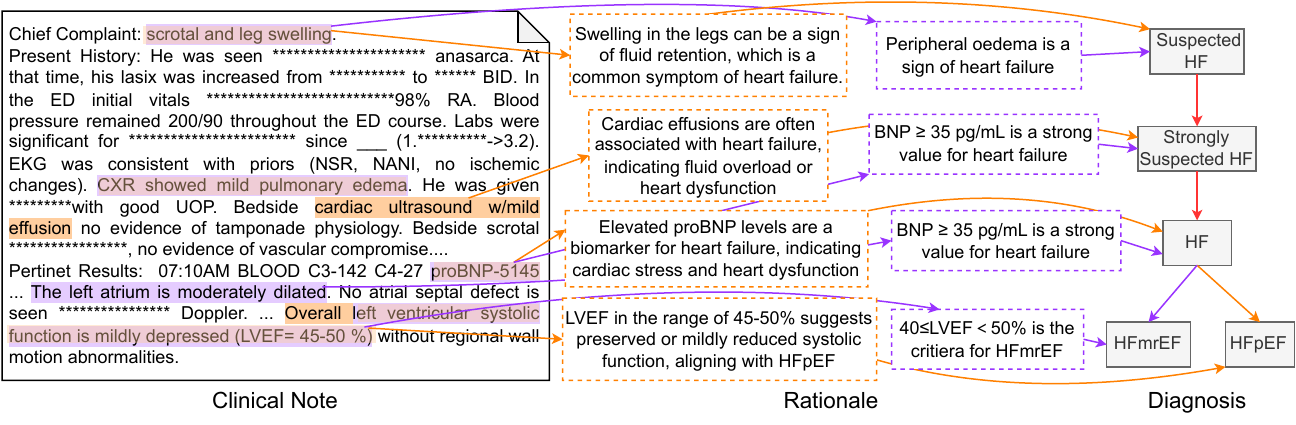}
\caption{An example prediction for a clinical note with PDD of \texttt{HF} by GPT-4}
\label{fig:HF}
\end{figure}

\begin{table}[t]
\caption{Prompt for $\mathcal{D}^{\star}$ setting.}
\centering
\resizebox{1\columnwidth}{!}{
\begin{tabular}{l}
\toprule
Input Prompt \\
\midrule
Suppose you are one of the greatest AI scientists and medical expert. Let us think step by step. \\
You will review a clinical 'Note' and your 'Response' is to diagnose the disease that the patient have for this admission. \\
All possible disease options are in a list structure: \color{red}{\{disease\_options\}}. \\
Note that you can only choose one disease from the disease options and directly output the origin name of that disease.\\
Now, start to complete your task.\\
Don't output any information other than your 'Response'.\\
'Note':\\
\color{red}{\{note\}}\\
Your 'Response':\\
\bottomrule
\end{tabular}
}
\label{open}
\end{table}

\begin{table}[t]
\caption{Prompt for no knowledge setting.}
\centering
\resizebox{1\columnwidth}{!}{
\begin{tabular}{l}
\toprule
Input Prompt \\
\midrule
Suppose you are one of the greatest AI scientists and medical expert. Let us think step by step. \\
You will review a clinical 'Note' and your 'Response' is to diagnose the disease that the patient have for this admission. \\
Note that you can only give one disease name and directly output the name of that "Disease".\\
Now, start to complete your task.\\
Don't output any information other than your 'Response'.\\
'Note':\\
\color{red}{\{note\}}\\
Your 'Response':\\
\bottomrule
\end{tabular}
}
\label{close}
\end{table}

\subsection{More Prediction Samples}

\ref{fig:HF} shows another sample generated by GPT-4. The ground-truth PDD of the input clinical note is and \texttt{Heart Failure (HF)}. In these figures, purple, orange, and red indicate explanations only in the ground truth, only in prediction, and common in both, respectively; therefore, red is a successful prediction of an explanation, while purple and orange are a false negative and false positive.

In Figure \ref{fig:HF}, the model's predictions do not align well with those of a human doctor. Key observations, such as the relationships between BNP and LVEF, are incorrectly identified, leading to a final misdiagnosis.

\subsection{Experiments for No Extra Knowledge}

We demonstrate the prompt used for $\mathcal{D}^{\star}$ and no knowledge settings in Table \ref{open} and Table \ref{close}, respectively. \{note\} is the text of whole clinical note and \{disease\_options\} in Table \ref{open} is the name of all leaf node $\mathcal{D}^{\star}$.

\subsection{Experimental Settings}
All experiments are implemented with a temperature value of 0. All close sourced models are implemented in a local server with 4 NVIDIA A100 GPU.

\section{Failed Attempts on DiReCT}
In this section, we discuss some unsuccessful attempts during the experiments.

\textbf{Extract observation from the whole clinical note.} We try to diagnose the disease and extract observation, and the corresponding rationale using the prompt shown in Table \ref{observation_fail1}. The \{note\} is offered by the whole content in the clinical note. We find that even though the model can make the correct diagnosis, only a few observations can be extracted (no more than 4), which decreases the completeness and faithfulness.

\begin{table}[t]
\caption{Prompt for extracting observation in one step.}
\centering
\resizebox{1\columnwidth}{!}{
\begin{tabular}{l}
\toprule
Input Prompt \\
\midrule
Suppose you are one of the greatest AI scientists and medical expert. Let us think step by step. \\
You will review a clinical 'Note', and your 'Response' is to diagnose the disease that the patient has for this admission.  \\
All possible disease options are in a list structure: \color{red}{\{disease\_options\}}. \\
Note that you can only choose one disease from the disease options and directly output the origin name of that disease.\\
Note that you also need to extract original text as confidence "Observations" that lead to the "Disease" you selected. \\
Note that you should extract all necessary "Observation".\\
Note that you also need to briefly provide the "Reason" for your extraction.\\
Note that both "Observations" and "Reason" should be string.\\
Note that your "Response" should be a list structure as following \\
:[["Observation", "Reason", "Disease"], ......, ["Observation", "Reason", "Disease"]]\\
Now, start to complete your task.\\
Don't output any information other than your 'Response'.\\
'Note'\\
:\color{red}{\{note\}}\\
Your 'Response':\\
\bottomrule
\end{tabular}
}
\label{observation_fail1}
\end{table}

\begin{table}[t]
\caption{Comparison for using Iteration and Once for observation extraction.}
\centering
\resizebox{0.92\columnwidth}{!}{
\begin{tabular}{lcccccc}
\toprule
&\multicolumn{3}{c}{Iteration} & \multicolumn{3}{c}{Once}\\
\cmidrule(lr){2-4}  \cmidrule(lr){5-7} 
Models & $\textit{Obs}^\text{pre}$ & $\textit{Obs}^\text{rec}$ & $\textit{Obs}^\text{comp}$ & $\textit{Obs}^\text{pre}$ & $\textit{Obs}^\text{rec}$ & $\textit{Obs}^\text{comp}$ \\ 
\midrule
LLama3 70B  & 0.277$_{\pm \text{0.146}}$ & 0.537$_{\pm \text{0.192}}$ & 0.256$_{\pm \text{0.142}}$ & 0.325 $_{\pm \text{0.207}}$ & 0.324$_{\pm \text{0.147}}$ & 0.185$_{\pm \text{0.107}}$\\ 
GPT-4 turbo & 0.446$_{\pm \text{0.207}}$ & 0.491$_{\pm \text{0.180}}$ & 0.371$_{\pm \text{0.186}}$ & 0.567$_{\pm \text{0.268}}$ & 0.287$_{\pm \text{0.156}}$ & 0.244$_{\pm \text{0.147}}$\\ 
\bottomrule
\end{tabular}}
\label{difference_ob}
\end{table}

\begin{table}[t]
\caption{Prompt for End-to-End prediction.}
\centering
\resizebox{1\columnwidth}{!}{
\begin{tabular}{l}
\toprule
Input Prompt \\
\midrule
Suppose you are one of the greatest AI scientists and medical expert. Let us think step by step. \\
You will receive a list of "Observations" from a clinical "Note" for the diagnosis of stroke.  \\
Here is the diagnostic route of stroke in a tree structure: \\
-Suspected Stroke\\
\quad\quad\quad -Hemorrhagic Stroke\\
\quad\quad\quad -Ischemic Stroke\\
\color{blue}{Here are some premise for the diagnosis of this disease. You can refer them for your task. Premise are: \{premise\}} \\
Based on these "Observations", starting from the root disease, your target is to diagnose one of the leaf disease.\\
Note that you also required to provide the "Reason" for your reasoning.\\
Note that your "Response" should be a list structure as following\\
:[["Observation", "Reason", "Disease"], ......, ["Observation", "Reason", "Disease"]]\\
Note that if you can't find any "Observation" to support a disease option, your "Response" should be: None\\
Now, start to complete your task. \\
Note that you should not output any information other than your "Response".\\
"Observations":\\
\color{red}{\{observation\}}\\
Note that you should not output any information other than your "Response".\\
Your "Response":\\
\bottomrule
\end{tabular}
}
\label{observation_fail2}
\end{table}

We also conducted an experiment to demonstrate the differences between two methods of observation extraction. The "Iteration" method is the one used in our paper, while the "Once" method is the one-time extraction method shown in Table 9. Each method was implemented under the condition of using GPT-4 turbo and Llama3 70B with $\mathcal{G}$ as input and was evaluated based on the Completeness of Observations (Obs) metric. The results are presented in Table \ref{difference_ob}. We found that while the "Once" extraction method resulted in higher precision, it led to a significant drop in recall, severely impacting the final completeness metric. The "Once" method tends to capture fewer observations, which hinders the overall reasoning process.

\textbf{End-to-End prediction.} We also try to output the whole reasoning process in one step (without iteration) when given observations. We show our prompt in Table \ref{observation_fail2}. We find that using such a prompt model can not correctly recognize the relation between observation, rationale, and diagnosis.

\section{Ethical Considerations}
Utilizing real-world EHRs, even in a de-identified form, poses inherent risks to patient privacy. Therefore, it is essential to implement rigorous data protection and privacy measures to safeguard sensitive information, in accordance with regulations such as HIPAA. We strictly adhere to the Data Use Agreement of the MIMIC dataset, ensuring that the data is not shared with any third parties. All experiments are implement on a private server. The access to GPT is also a private version.

AI models are susceptible to replicating and even intensifying the biases inherent in their training data. These biases, if not addressed, can have profound implications, particularly in sensitive domains such as healthcare. Unconscious biases in healthcare systems can result in significant disparities in the quality of care and health outcomes among different demographic groups. Therefore, it is imperative to rigorously examine AI models for potential biases and implement robust mechanisms for ongoing monitoring and evaluation. This involves analyzing the model's performance across various demographic groups, identifying any disparities, and making necessary adjustments to ensure equitable treatment for all. Continual vigilance and proactive measures are essential to mitigate the risk of biased decision-making and to uphold the principles of fairness and justice in AI-driven healthcare solutions.

\newpage
\qquad
\newpage


\bibliography{citations}

\end{document}